# Towards a Unified System of Representation for Continuity and Discontinuity in Natural Language

**Ratna Kandala**[1]
n038k926@ku.edu
Department of Psychology,
University of Kansas, USA

**Prakash Mondal**[2]
prakashmondal@la.iith.ac.in
Department of Liberal Arts,
Indian Institute of Technology Hyderabad, India

## Abstract

Syntactic discontinuity is a grammatical phenomenon in which a constituent is split into more than one part because of the insertion of an element which is not part of the constituent. This is observed in many languages across the world such as Turkish, Russian, Japanese, Warlpiri, Navajo, Hopi, Dyirbal, Yidiny etc. Different formalisms/frameworks in current linguistic theory approach the problem of discontinuous structures in different ways. Each framework/formalism has widely been viewed as an independent and non-converging system of analysis. In this paper, we propose a unified system of representation for both continuity and discontinuity in structures of natural languages by taking into account three formalisms, in particular, Phrase Structure Grammar (PSG) for its widely used notion of constituency, Dependency Grammar (DG) for its head-dependent relations, and Categorial Grammar (CG) for its focus on functor-argument relations. We attempt to show that discontinuous expressions as well as continuous structures can be analysed through a unified mathematical derivation incorporating the representations of linguistic structure in these three grammar formalisms.

**Keywords**: constituency, discontinuity, dependency, functor-argument relations, unified representation

## 1. Introduction

Analysing sentences as structured entities is a very old idea which contributed to the development of pure logic. The basic notion of constituency was founded on the semantic distribution between predicates (properties and the relations between entities) and arguments (participants in the predicate relations which complete the meaning of a predicate) and can be traced back as far as Apollonius Dyscolus (200 AD) and Aristotle (Carnie 2010). During the 20$^{th}$ century (since 1950s) the research program of 'generative linguistics' emerged as an influential study of language. Generative linguistics/Generative grammar is a field of study originating from the work of Noam Chomsky that attempts to discover the nature of the human language faculty and achieve the goal by focusing on the purely internal component of the language – syntax. Within the generative tradition, the first model of syntax presented by Chomsky in 'Syntactic Structures' in 1957 -transformational syntax- is the best known. Later developments of this model are known by various names such as the Standard Theory, the Extended Standard Theory, The Revised Extended Standard Theory, Government/Binding theory and the Minimalist Program. They all exhibit certain common characteristics: all syntactic representations are immediate-constituent structures (ICS); they are represented using trees; the grammatical functions/relations (subject, object) are derived from the constituent structure in the trees; the configuration of the subject is higher and external when compared to the object. Certain operations called transformations (hence transformational grammar) on an existing constituent structure change it into a similar, but not identical, constituent structure called the 'surface structure'. Therefore, this approach is also known as a derivational approach since the most remarkable feature is the set of consecutive representations of a grammatically well-formed sentence. The idea of ICS dates back to the heyday of American structuralism and this notion was developed further by Leonard Bloomfield in American structuralism and later by Charles Hockett. Constituents under the ICS analysis had a distributional basis too, as this came to be developed in the work of Harris (1951, 263-264) who proposed that sequences of morphemes occurring in the same distributional context would count as identical constituent types. In essence, the ICS analysis splits sentences into constituents by examining the proximity of modification relations among the words. For instance, the constituents of *The cat chased the rat* would be *the, cat, chased, rat, the cat, chased the rat, the rat, chase, ed* (Carnie 2010: 70). A ‘phrase’ is thus understood as a group of words which behave syntactically (and semantically) as a unit/constituent. When such words are adjacent to each other, they are easily grouped under a single category in the corresponding phrase structure. An essential element of phrase structure is that each phrase contains a head and is a projection from the head of the phrase. Each lexical category projects to a phrasal category. This is roughly the Head Constraint (the widely held assumption in classical generative grammar) on argument structure which says that the head X of a syntactic phrase XP expresses (i) a semantic function F and (ii) its obligatory (YP) and optional (ZP) syntactic arguments that express the semantic argument/s (of F) W and V, corresponding to YP and ZP respectively (Jackendoff 2002, 145). In other words, all the syntactic arguments and adjuncts are located within the same phrase and they express the semantic arguments and modifiers of the head word.

Such an understanding is applicable to languages such as English which is indicative of ‘continuity’. Languages such as Turkish, Russian, Japanese, Croatian, German, Hindi, Tamil, Warlpiri (a native Australian language), and also American Southwest languages like Navajo, Hopi, Dyirbal, Yidiny, etc. exhibit an interesting phenomenon in which even when the elements of a phrase are not adjacent to each other (which is indicative of ‘discontinuity’), they form grammatical sentences. In other words, syntactic discontinuity is a grammatical phenomenon in which a constituent of a sentence is split into two (or more) parts because of the insertion of an element which is not a part of the constituent. The evidence for discontinuity is frequently found in languages with relatively free word-order such as the ones mentioned earlier. Discontinuity is also observed in fixed word-order languages like English in phenomena such as topicalisation, relative clauses, long-distance dependencies (Eg: the relation between a *wh-*/topicalized phrase and its trace in a sentence), indirect questions, parenthetical expressions (such as 'She knows, and I'm sure you do too, all that we have found out'),and also in cases when an NP argument is embedded within a PP as in [$_{S}$ *Max talked* [$_{PP}$ *to* [$_{NP}$ *Sam*]$_{NP}$ ]$_{PP}$ [$_{PP}$ *about* [$_{NP}$ *the picture* ]$_{NP}$ ]$_{PP}$ ]$_{S}$ and raising predicates, as in *Georg seems [to like ice cream].* 'Georg' is the semantic argument, not the subject, of the verb 'like' and the semantic argument is in a position higher than the verb's subject position, thereby violating the Head Constraint (Hudson 1994, Van Valin 2001, Jackendoff 2002, Citko 2011). While not all of these phenomena may involve the insertion of an intervening element/expression not part of the constituent expression, they appear to break the continuity of the constituent structure. Further, Ott (forthcoming) argues that sentences such as *Daisy has brought, Jasmine tells me, a nice coat today* pose a serious challenge to the constituency-based organization of phrase structure in that the interpolated parenthetical expression *Jasmine tells me* disrupts the overall arrangement of component parts, even though these parts show signs of internal organization. In any case, regardless of whether or not such cases are limited owing to the rigid word order patterns within and across languages, they deserve attention and explanation.

Importantly, other approaches to syntax, rejecting some or all of these assumptions of the derivational model were Generalized Phrase Structure Grammar (Gazdar 1983; Gazdar et al. 1985), Pollard and Sag's Head-driven Phrase Structure Grammar (1987), Bresnan's Lexical Functional Grammar (1982), Dependency Grammar (DG) (Tesniére 1959), Categorial

Grammar (CG) (Ajdukiewicz 1935, Steedman 1992, Steedman 2014), Dependency Constituency Grammar (Barry and Pickering 1990). One of the main differences between these types of grammar formalisms lies in the manner in which they approach discontinuous constituents. PSG has rules and analyses syntactic structures only in terms of constituents/phrases, making it well-nigh impossible to account for discontinuous constituents (for example, tangled trees could be an exception, see section 5.3). DG and related formalisms typically eliminate the idea of a phrase/constituent and accommodate discontinuous constituents by doing away with the idea of constituents.

However, an integrative system of representation which accounts for the two systems of representation- rigid and flexible constituency incorporating the idea of a phrase structure, and head-dependent and functor-argument relations- has not been suggested so far. Is there a way in which the two systems of representation can be integrated? Are they two different systems of representation that cannot really exist as one system of representation but can only have parallel descriptions? These are the questions this paper attempts to address, and it formulates a unified system of representation through a series of mathematical derivations in order to integrate representations of both continuous and discontinuous linguistic structures with a special focus on three *context-free grammar* (CFG) formalisms: PSG, DG and CG. This is significant in the current debates on the question of whether constituency-based grammar formalisms can be translated smoothly into non-constituency formalisms (such as DG) or vice versa (see for discussion, Müller 2019).

Section 2 introduces the understanding of a constituent with an indication of its significance in linguistic theory. The evidence for discontinuity is discussed in section 3. Section 4 discusses different approaches towards discontinuity, namely the three formalisms considered for this study, in order to show how they square up to the problem of discontinuity. Section 5 discusses the different solutions proposed so far in the literature to tackle the problem of discontinuity and how well they fare in structurally accommodating discontinuous structures of different kinds. Section 6 illustrates with derivations (bidirectional) the *correspondence principle* that can help unify the representations in the three formalisms. The overarching goal is to demonstrate how continuity and discontinuity in natural language can be accommodated in one single system of representation. Finally, section 7 dwells on the implications of the current work for linguistic theory and also offers some concluding remarks.

## 2. The Notion of a Constituent

'Constituent' is an umbrella term used to refer to a word or a phrase or a sentence. Constituents are represented by either using tree diagrams or using the bracketing method. When two smaller units (considered from the level of words) combine to form a larger unit, a hierarchy is established between them, and the larger unit is referred to as a 'syntagm' (though the usage of this term has become redundant) (Matthews 1981). Thus, a constituent structure is well captured as a hierarchical tree rather than a mere concatenation. This structure is well in tune with our intuitions, when we encounter a meaningful sentence, as to which words go together and which do not. Hence a syntagm consists of units called ‘constituents’ enclosed within, and of these, a direct relationship is established by what are referred to as the ‘immediate constituents’. The ‘root’ node is the highest node referring to the entire clause/phrase. Syntagms are classified as NP/PP/AP/VP based on their constituents and the constituent acting as the head of the corresponding phrase. Thus a ‘phrase’ and a ‘syntagm’ are synonymous with each other. An advantage of such an understanding is that it allows us to account for recursion (the schema XP repeats itself within the very same XP scheme where X is the head) where a syntagm of one class can contain another syntagm of the same/different class indefinitely. This is *arguably* a distinctive phenomenon of human language in its ability to permit unboundedly long expressions/strings from a finite number of elements primarily because it allows for numerous creative thoughts to be expressed by humans unlike other animals. Consider *a girlfriend of the man from the team on the street*. The rewriting rules are- Rule 1: NP → Det N PP and Rule 2: PP → P NP. Here one NP is embedded inside another NP via a PP (NP → Det N P NP). Thus, the length of the expression can become infinitely long (Everaert et al. 2015). However, Pullum & Scholz (2010) have challenged the argument that all the possible expressions in any language form an infinite set[1]. That is because this postulation about infinitude is based on the assumption that every expression length has a successor and that no two expression lengths share a successor (just like the mathematical induction that every integer *n* has a successor *n+1*). This assumption is not established. Nor is it tenable, precisely because both *She likes to buy stuff* <u>*that is available in the nearby market*</u> and *He often gets to see almost everything* <u>*that is available in the nearby market*</u> share the same underlined expression as the successor (see for related discussion, Anonymous 2014). In any case, the goal here is not to capture this debate; rather, it is merely to indicate that alternative interpretations of unbounded expressions in constituency-based analyses exist in the literature.

## 2.1 The significance of constituents

Analysing sentences in terms of constituents captures the essential idea that a group of words is not a linear concatenation, but considered somewhat similar to a kind of combination which is not commutative. This is captured by the hierarchical structure of a tree diagram. For instance, a sentence such as *Ronny types very fast* can be decomposed into two major constituents *Ronny* and *types very fast*, but the combination in the grouping (*types (very fast))* is not commutative (e.g. neither *very fast types* nor *types fast very* is well-formed). This kind of closer structural relation among some words, unlike the case of abstract numbers, is also linked to our linguistic intuition and, interestingly, both linearity and agreement relations (such as subject-verb agreement) are accommodated in a hierarchical structure of a tree. This is clearly evident in the case of non-local dependencies and subject-auxiliary inversions (SAI) (Matthews 1981, Leffel and Bouchard 1991, Moro 2008, Carnie 2010, Everaert et al. 2015). The essential goal here is to spell out the significance of constituency via agreement relations as defined on hierarchical structures in constituency-based frameworks (see for an alternative view, Bod 2009).

[1] Pullum & Scholz (2010, 121) have also gone further to establish that recursion does not necessarily generate an infinite set by considering a simple example of a context-sensitive grammar. The rules for this grammar are S → NP VP; VP → VP VP ; NP → *They* ; VP → *came / They_ ;* VP → *running / They came_*. The second rule VP → VP VP is a recursive rule and yet generates only two strings: *They came* and *They came running*. Hence recursion does not guarantee the possibility of having an infinite set.

## 3. Evidence for Discontinuity: Characterization of Non-configurational Languages, Lexical Structure (LS), and Phrase Structure (PS)

The work by Hale (1982, 1983) on Australian native languages such as Warlpiri etc., American Southwest languages like Navajo, Hopi and by Dixon (1972, 1977) on Dyirbal and Yidiny has provided a wealth of evidence for discontinuity in natural languages. This is because the syntactic nature of these languages is not the same as that of more familiar languages which admit of analyses in terms of phrase structure constituency (the structure of a clause, configurations of NPs and VPs), subordination, *wh*-movement and extraposition (McGregor 1992, Nordlinger 2014). Accordingly, these (the former) are referred to as non-configurational languages since they permit discontinuous constituents.

Hale (1982, 1983) associated three key properties with ‘non-configurational languages’: (i) free word order (ii) the use of syntactically discontinuous expressions, and (iii) the extensive use of null anaphora (an argument (such as subject and object) that is not represented by an overt nominal expression in the phrase structure). According to him, the difference between fixed word order languages (like English) and free word order languages (like Warlpiri) lies *not* in the phrase structure, but in the nature of the relationship between the 'lexical structure (LS)' (predicate-argument structure) and the 'phrase structure (PS)'. He defined the 'logical form (LF)' of a clause by the relation of the LS to the PS where a nominal in PS is linked with an argument in LS. To illustrate this point, let us consider the examples (1-3) from Warlpiri:

(1) *Panti-rni* ***ka*** *ngarrka-ngku* *wawirri*

spear-NON-PAST AUX man-ERG kangaroo

'The man is spearing the kangaroo'

(2) *Ngarrka-ngku* ***ka*** *wawirri* *panti-rni*

man-ERG AUX kangaroo spear-NON-PAST

'The man is spearing the kangaroo'

(3) *Wawirri* ***ka*** *panti-rni* *ngarrka-ngku*

kangaroo AUX spear-NON-PAST man-ERG

'The man is spearing the kangaroo' (Hale 1983:6)

[NON-PAST=non-past tense; AUX=auxiliary; ERG=ergative case marking]

In (1-3) the ergative nominal *ngarrka-ngku* in the phrase structure (PS) corresponds to the subject argument in the lexical structure (LS) of the verb *panti-rni*. Similarly, the absolutive nominal *wawirri* in the PS corresponds to the object argument in the LS. Hence even (2) and (3) are grammatical sentences with the same meaning as (1) in Warlpiri (the only requirement in Warlpiri is that ‘AUX’ should be in the ‘second position’, that is, in Wackernagel’s position as per Wackernagel’s law stating that unstressed forms including auxiliary forms or clitics are to be placed in the second position of a clause, as marked in bold).

According to Hale, in configurational languages, for each argument in the LS there must be a corresponding constituent in the PS, and this LS-PS relation is therefore identical. This, in a way compels the phrasal syntax to exhibit the NP-VP organization of clauses. However, in non-configurational languages, such one-to-one correspondence between LS arguments and PS nominal expressions does not exist. Accordingly, this gives rise to the manifestation of the three key properties of non-configurational languages mentioned earlier, that is, free word order, the use of syntactically discontinuous expressions and a null anaphora.

### 3.1 NP constituency

Nouns and adjectives in English are considered to be different lexical categories given their different morphological and syntactic functions. In many Australian languages, they show the same morphological and syntactic properties; hence it's difficult to distinguish them as separate word classes. This also questions the universality of AP. It is well known that expressions denoting quality (such as adjectives) typically function as modifiers, while pronouns, proper nouns, and common nouns function as arguments. However, in specific contexts, nouns can also function as modifiers, akin to adjectives. Consider the following discontinuous sentence (4) from Warlpiri (Nordlinger 2014, 238):

(4) *Maliki-rli=ji* *yarlku-rnu* *wiri-ngki*

dog-ERG = 1.OBJ bite-PAST big-ERG

'The/a big dog bit me'; 'The/a dog bit me, a big one'

[PAST=past tense; OBJ=object]

There are two interpretations possible (Here both *wiringki* and *malikirli* are nominals in Warlpiri). First, we may consider the case where *maliki* is the head and *wiri-* is the modifier. This means 'The/a big dog bit me.' In other words, *wiringki* acts as a modifier for the head *malikirli* (Hale called this 'merged' interpretation). Second, we may consider the case where either *wiringki* or *malikirli* can be the head and the relation between them can be restrictive or non-restrictive modification. This gives rise to the other interpretation 'The/a dog bit me, a big one', (Hale called this 'unmerged' interpretation).

In addition to this, although it is not unusual to find nominals functioning as predicates in English and European languages (as in copula constructions), they are very frequently found in Australian languages, as can be seen in the Warlpiri sentence (5).

(5) *Ngarrka* *ka-rna* *nya-nyi*

man (ABS) PRS-1SG.SBJ see-NON-PAST

[ABS=Absolutive case marking; SBJ=subject; PRS=present tense; 1SG=first person singular]

a) 'I see the/a man' (argumental reading)

b) 'I see him (as a man / and he is a man)' (predicative reading)

Accordingly, Hale argued that two readings are available to nominals in Warlpiri: 1. Argumental reading 2. Predicative reading (this is possible because the nominals function as secondary predicates). Here, the widely adopted template of an NP constituent, used in many Australian languages, has been proposed by McGregor (1992): (Deictic) (Quantifier) (Classifier) Entity (Qualifier). Each of these functions may be realised by different subclasses of nominals. Thus, the main idea here is to show that a nominal constituent is not same thing universally and variation in functions of the NP constituent gives rise to certain kinds of discontinuity.

**3.2 VP constituency**

As discussed earlier, Hale (1982, 1983) has shown that Warlpiri and Wambaya require the 'AUX' in the second 'Wackernagel' position. Accordingly, the auxiliary follows many phrasal constituents like complex NPs, subordinate clauses, but not necessarily a VP. This piece of evidence questions the universality of a VP because the Aux cannot follow the sequence of verb+object, which suggests that they cannot form a VP. Also, the standard VP constituency tests fail in Warlpiri (Nordlinger 2014), as discussed previously.

Languages such as Kayardild (spoken on the South Wellesley Islands, Queensland, Australia) are free word order languages at the clausal level but do not freely allow discontinuous NP constituents. Mandarin allows NP arguments to be dropped despite not being a free word order language. Given these variations among the world languages, questions arise about the validity of the widely accepted notion of a phrase/constituent which is not universally present in all the languages, and this offers the motivation for coming up with a more adequate system of representation for accommodating these morpho-syntactic differences.

## 4. The Three Grammar Formalisms: PSG, DG, CG

We shall provide a brief introduction to PSG, DG and CG and then illustrate the approach of each of the three formalisms towards discontinuity.

### 4.1 PSG's account of discontinuity

The PSG analysis of the Warlpiri sentences in (1-3) is illustrated below with each of the sentences repeated without the glosses. As can be clearly seen in Figure 1, the meaning remains the same but each of the sentences in (1-3) has a different tree structure.

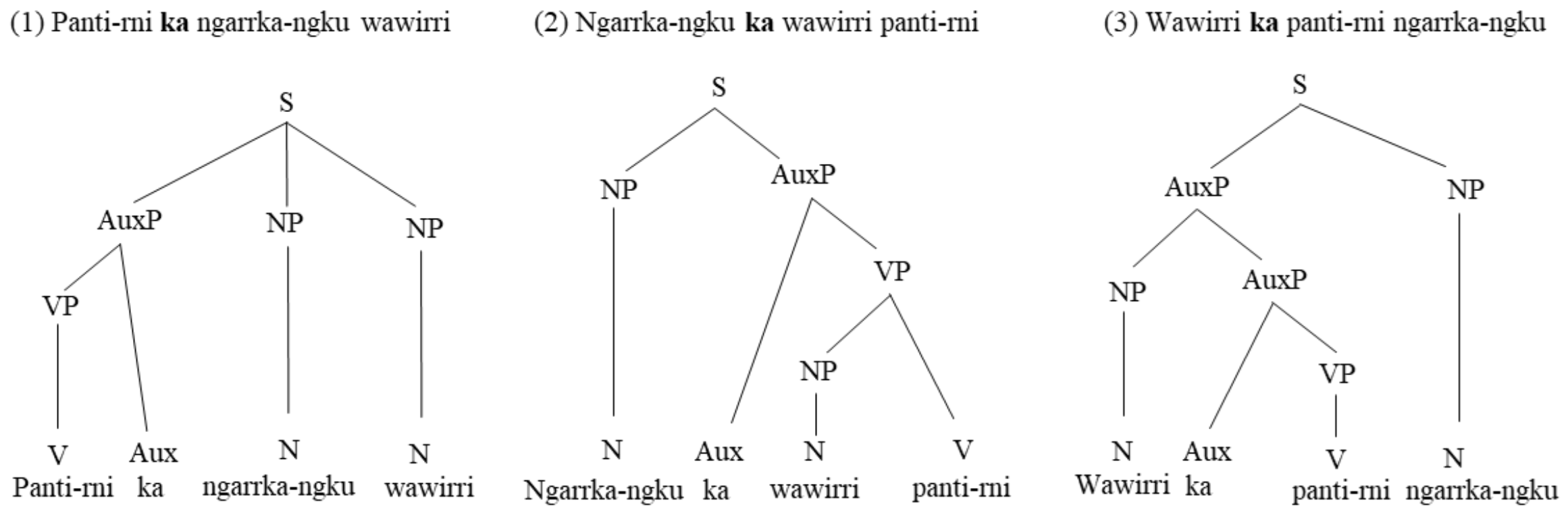

Fig 1. PSG trees for examples (1-3)

**4.2 DG's account of discontinuity**

The origins of this grammar formalism can be traced back to Panini. However, an attempt to build a complete theory of grammar based on dependency was made by the French linguist Lucien Tesnière (1959). He was one of the pioneers in demonstrating the potential of dependency grammar in capturing the similarities and differences across languages. DG is a descriptive tradition in linguistics which assumes that syntactic structure consists of binary asymmetrical relations (this refers to the mapping of the elements of a sentence (words/phrases/syntactic relations) to the actual words of a sentence). These are referred to as dependency relations or dependencies between words. The relation is asymmetric because one word which is understood as the dependent complements/modifies the other words that are regarded as heads (a one-way dependency). Sometimes these are also referred to as the modifier (dependent) and the governor (head). The basic idea is that all except one word in a sentence depends on another word. The 'root' (independent/main/central element) refers to the one word in a sentence that does not depend on any other word in the sentence. 'Grammatical functions' (syntactic as well as semantic) motivate the dependencies. A dependent Y depends on a head X when Y is (usually) optional with respect to X, where X selects Y, and/or Y agrees or is governed by X, and/or the linear position of Y is with reference to that of X (de Marneffe and Nivre 2019, 203). It may be noted that not all these conditions may hold in tandem in a given case of dependency relation, because when, for example, X selects Y or Y is governed by X, we cannot say that Y is optional with respect to X. In any case, the machinery of DG usually does away with the PSG notion of a constituent and focuses on the concept of 'valency' directly borrowed from chemistry, although a sub-graph based (or sub-tree based) notion of constituents is adopted by some DG theorists (Hudson 1984, 92; Miller 2000, 22). The concept of valency is important in DG because different predicates take different numbers of arguments. For example, a transitive verb takes two arguments, meaning two dependents. Such kind of representations may include information about the surface order of the string being analysed (Hays 1964, Debusmann 2000, de Marneffe and Nivre 2019).

*A formal definition of DG*: A DG can be specified by a 4-tuple: DG = $<V_N, V_T, D, R>$ where $V_N$ is the set of non-terminal vocabulary items (referring to the categories), $V_T$ is the set of terminal vocabulary items (referring to the actual words), D is the set of dependency rules and R is the initial symbol at the root of the tree (see also. There are two important rules in D:

Rule 1: $I(D_1,\ldots,D_m * D_i,\ldots,D_k)$ (i, m, k $\geq 0$ ; not always i=m=k)

Rule 2: I (*)

'I' indicates the presence of only one independent category; $D_1,\ldots D_m$ represents the dependent categories on the left of the root word; $D_i,\ldots.D_k$ represents the dependent categories on the right of the root word; ‘*’ indicates the location of 'I' in the string of dependent words.

Unlike phrase-structure trees, dependency trees are not actually sensitive to the word order of a sentence and, therefore, this is an advantage for an analysis of free-word order languages such as Japanese, Warlpiri, Korean, Finnish, German, Latin, Russian, Kalam (spoken in Papua New Guinea) etc. For example, there is one DG graph for the English sentence *The child rides the bicycle* and it’s topicalized (discontinuous) form *the bicycle, the child rides*. Another form of flexibility is non-projectivity that may arise from word order variation. Thus, for example, a sentence such as *A decision is pending on this matter* will involve crossing between the edge from *decision* to *on* and the projection lines that may be drawn vertically from *is* and *pending*. Likewise, a sentence like *Spaghetti, Mary wants to eat* *today* will also involve a dependency relation between *eat* and *spaghetti* crossing the intervening sequence of *Mary wants*, since each word from *Mary wants* will have a vertical projection line (see Debusmann 2006, 17; see also Hudson 1994). This is advantageous because this "projectivity restriction" is optional in DG unlike the obligatory restriction on continuity in PSG (Debusmann 2006: 17). None of the rules framed above as part of the formal definition bans non-projectivity.

Crucially, it may be possible to arrange the dependents in such a manner that their surface order can be shown (Debusmann 2000, 3), even though dependency relations have no intrinsic linear order in Tesnière's (1959) original formulation. Also, Hudson's (1984) Word Grammar anchored in DG heavily relies on linear order. In any case, the underlying principle is that DG relations are not *intrinsically* sensitive to word order, and this can be illustrated by the fact that there is a single dependency tree corresponding to examples (1-3), as seen in Figure 2. It may be observed that we can define a dependency valuation function (δ) that takes a node as an input and returns a real value as an output (see Levelt 2008: III:51). If A~B (meaning that A is dependent on B), then δ(A)>δ(B). The real value is set to 0 at the top of the tree, but we can start from 1 at the top of the tree. This function will be useful for recoding CG functor-argument relations in terms of dependency relations, as will be shown in Section 6.

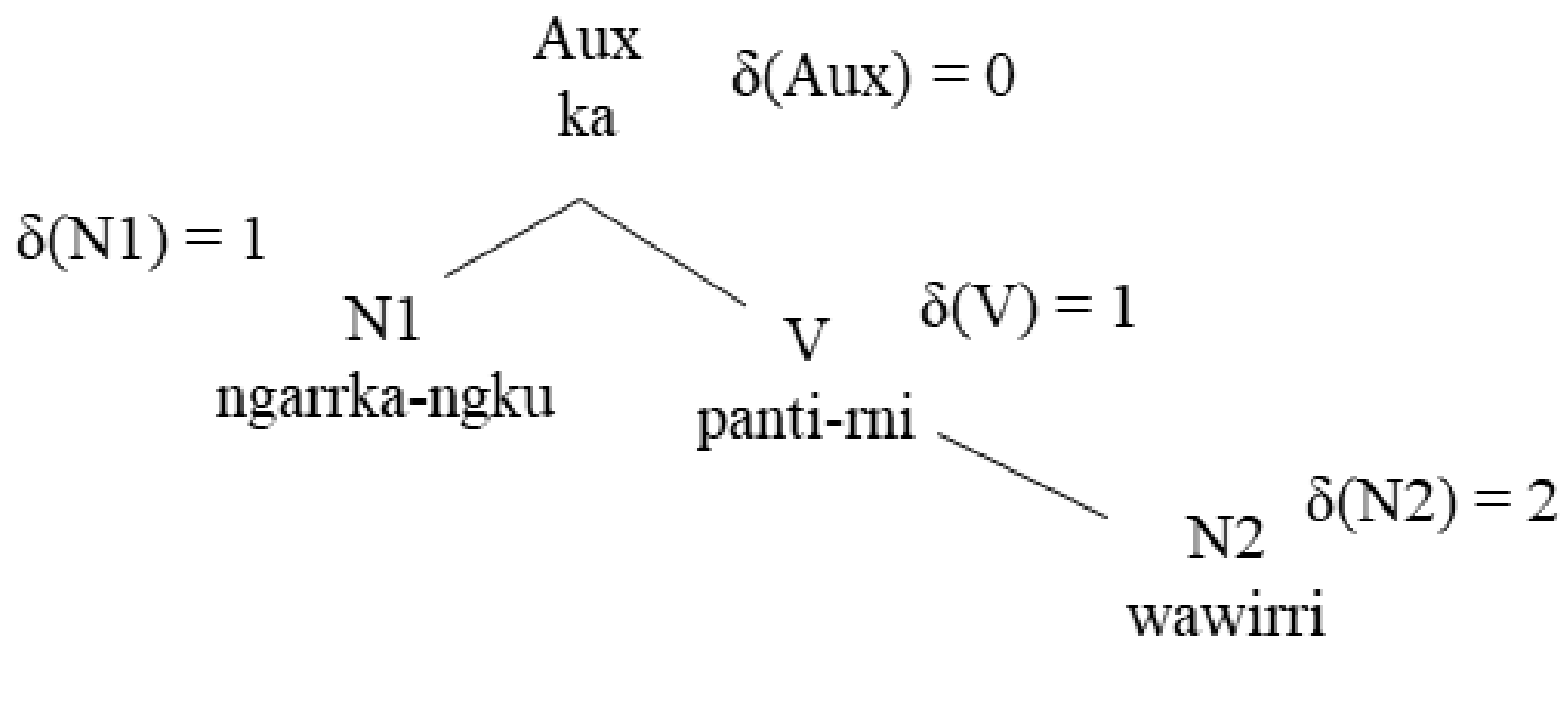

Fig 2. A dependency graph of the discontinuous Warlpiri sentences (1-3)

Thus, there are certain differences of dependency graphs over tree diagrams. A major difference is there is one-to-one mapping between the number of nodes in the tree and the number of words in the sentence in a dependency graph. However, departures from this convention are not hard to find, although we do not intend to claim that these departures are an intrinsic part of the DG formalism. In fact, some dependency analyses neutralise the one-to-one mapping condition due to having to show the category-to-word mapping (see Anderson 2011, 95-96; Hudson 2018, 97), or the speech act of a sentence (see Levelt 2008, II:136), or the root value of a dependency graph for dependency algorithms (see Gebhardt, Nederhof and Vogler 2017, 466). Since nothing substantial depends on this specific convention for our purpose, we would not make much of one-to-one mapping in our dependency analyses in this paper. In a phrase structure tree, many-to-one mapping typically exists (the number of nodes is always greater than the number of words). This generalization notwithstanding, constituency-based grammars (especially bare phrase structure) may also be found to be gravitating towards the minimization of the number of nodes, thereby approximating one-to-one mapping (see Osborne, Putnam and Gross 2011). Another crucial difference between the two is the manner in which the heads are identified. In phrase structure trees, the labels are marked on the nodes to indicate the head of the phrase. In a dependency tree, the head of a given word is the immediately dominating word and the dependent is a word dominated by the head word. This is inherently identified in the structure of the tree (Osborne 2019).

### 4.3 CG's account of discontinuity

Categorial Grammar is a context-free grammar formalism first explicitly defined by the Polish logician Kazimierz Ajdukiewicz (1935). The notion of 'category' is crucial to this formalism. All grammatical constituents and all lexical items in particular are associated with a ‘category’, hence referred to as ‘categorial’ grammar. This ‘category’ defines the potential of a lexical item to combine with other constituents/expressions to yield compound constituents. The widely used ‘slash’ notations for directional categories were pioneered by Bar-Hillel (1953) and Lambek (1958). Lambek’s notation uses a forward slash ‘/’ to indicate an argument on the right and a backward slash ‘\’ to indicate an argument on the left. A category such as X/Y represents a function looking for an argument of category Y on its right and resulting in the category X. The same notation will be used for the current analysis. It permits words of a given sentence to be combined in a number of different ways. Words are assigned a category in terms of N and S, based on their combining properties (Steedman 1992, Steedman 2014). It needs to be emphasized that for all our CG analyses we have adopted the standard Lambek notation of functor-argument relations (by using only the categories N and S).

Constituents are characterized as functions and arguments, both syntactically and semantically. The primitive notion of a head does not exist in CG, but the notion of a semantic/syntactic functor is often considered to be head-like (Barry and Pickering 1993). Thus, the idea of a flexible (discontinuous) constituent is very effortlessly explained in this formalism. The CG derivation of a sentence proceeds as per the constituent structure in phrase structure formalisms, as seen in Figure 3 for the following Malayalam sentence (6) (Falk 2001,19).

(6) *Kaṇṭu* *kuṭṭi* *aanaye*.

saw child.NOM elephant.ACC

‘The child saw the elephant.’

[NOM=nominative case marking; ACC=accusative case marking]

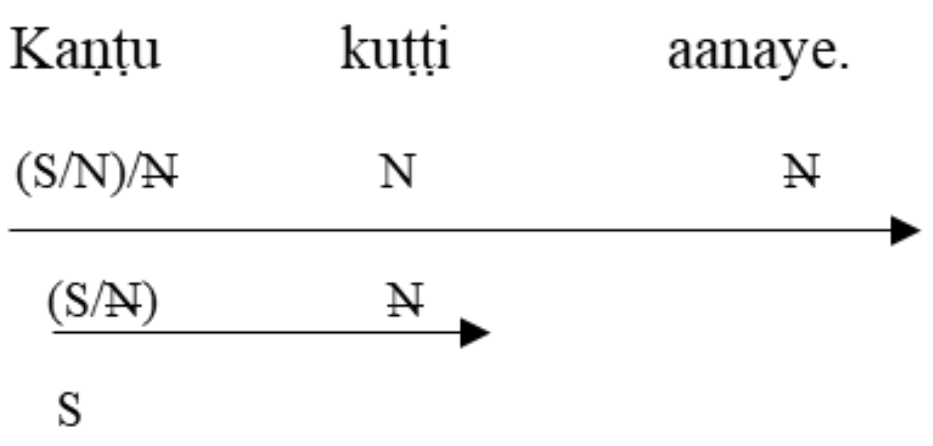

Fig 3. A CG derivation

*A formal definition of CG:* A CG can be defined by a 4-tuple: CG = <V,C,R,F> where V is the set of all lexical items in a language, C is the set of primitive categories ({N,S}), R is the set of functional composition rules for the generation of categories for lexical items and F is a function that maps each lexical item (LI) to its set of categories (each element of V is mapped to its corresponding element(s) which can be a set of primitive/atomic categories from the set C and/or categories derived by means of R), whose form is: F(LI) = $\{C_1 ,\ldots,C_n\}$. Thus, CG, unlike PSG, holds a significant advantage in its capacity to explain discontinuous phenomena at the morphological level (see Morrill 1995).

Also, CG accounts for certain kinds of discontinuity by using Type-Raising rules to allow arguments in coordination sentences such as *I cooked, and she ate, the rice*. Here *I* and *cooked* need to combine to yield the output (S/N) and likewise *she* and *ate* will yield (S/N). Once *I cooked* and *she ate* are combined together via *and*, all we will be left with is (S/N), allowing *the rice* to be finally combined/composed (Steedman 2014, 676). This maintains the assumption that constituents that can coordinate are formed without deletion or movement in such sentences and led to the proposal of the Forward Type-Raising rule: Y => X/(Y\X) where the category of the NPs *I* and *she* is type raised from N to S/(N\S) (a briefer notation). It may be argued that the wrapping operation (discussed below) will be more relevant to most of the examples considered and discussed in this manuscript, although type-shifting may be required in other alternative cases. However, we leave this matter open for discussion.

## 4.4 A comparison of the three grammar formalisms

In this section, we compare the three formalisms by considering a discontinuous sentence in Kwakwala, a Wakashan language of British Columbia with verb-subject-object (VSO) word order as seen in (7).

(7) *Yəlk$^{w}$əmas-Ø-ida* *bəg$^{w}$anəma-x̣-a* *ὠats’i-s-a* *g$^{w}$ax̣ƛ̓ux̣$^{w}$*

hurt-NOM-DEM man-ACC-the dog-INST-the stick

‘The man hurt the dog with the stick.’ (Van Valin 2001:120)

[DEM=demonstrative; INST=instrumental case marking]

The most striking feature of this language, as seen in (7), is that the nominative case marker (*-Ø-*) and a demonstrative *(-ida*) attached to the verb *Yəlk$^{w}$əmas-* actually qualify the following N *bəg$^{w}$anəma-*.

(7a) A phrase structure analysis:

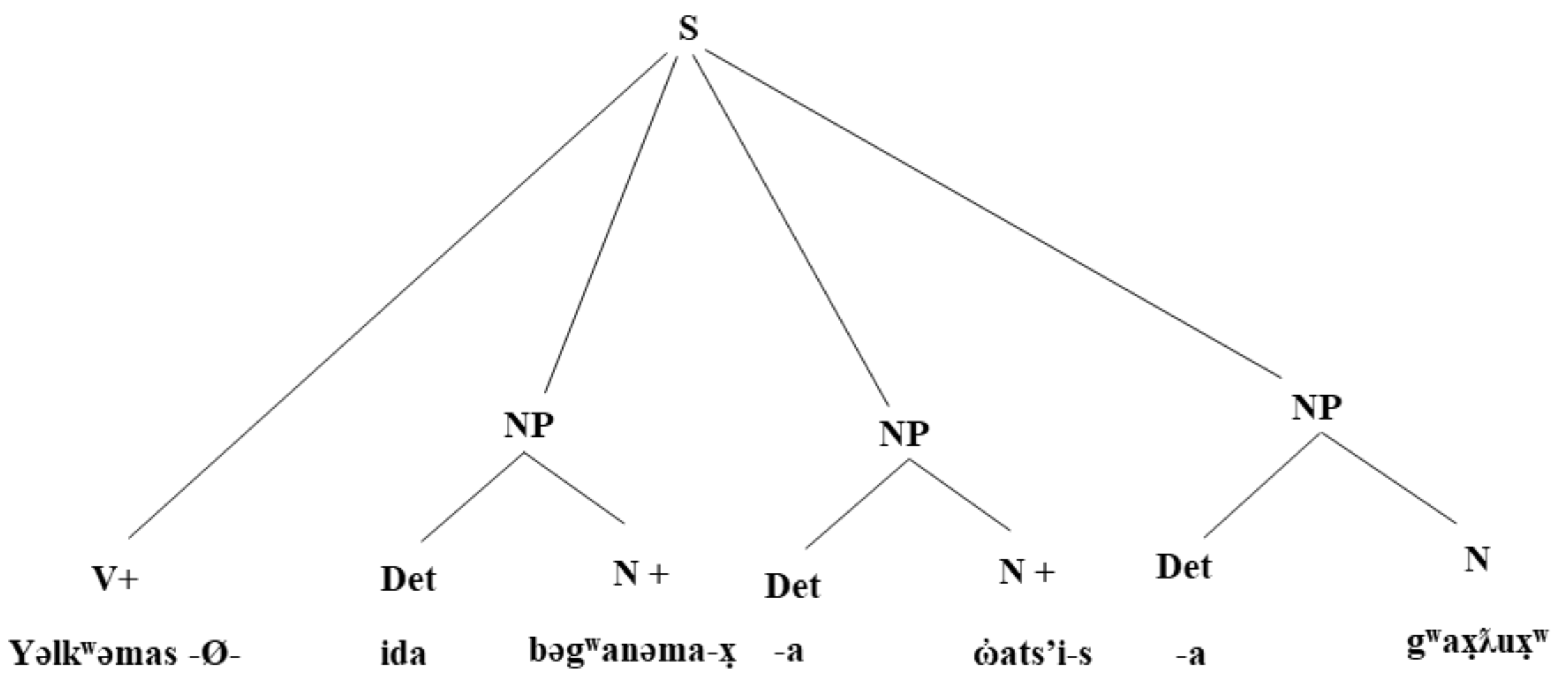

Fig 4. A PSG tree of (7)

Note: In figure 4, the constituent *Yəlk$^{w}$əmas-* is analysed as a V, not as a VP, mainly because the organization of the VP as a continuous unit is distorted by the presence of the subject NP between the V and the object NP (Van Valin 2001).

(7b) A dependency analysis:

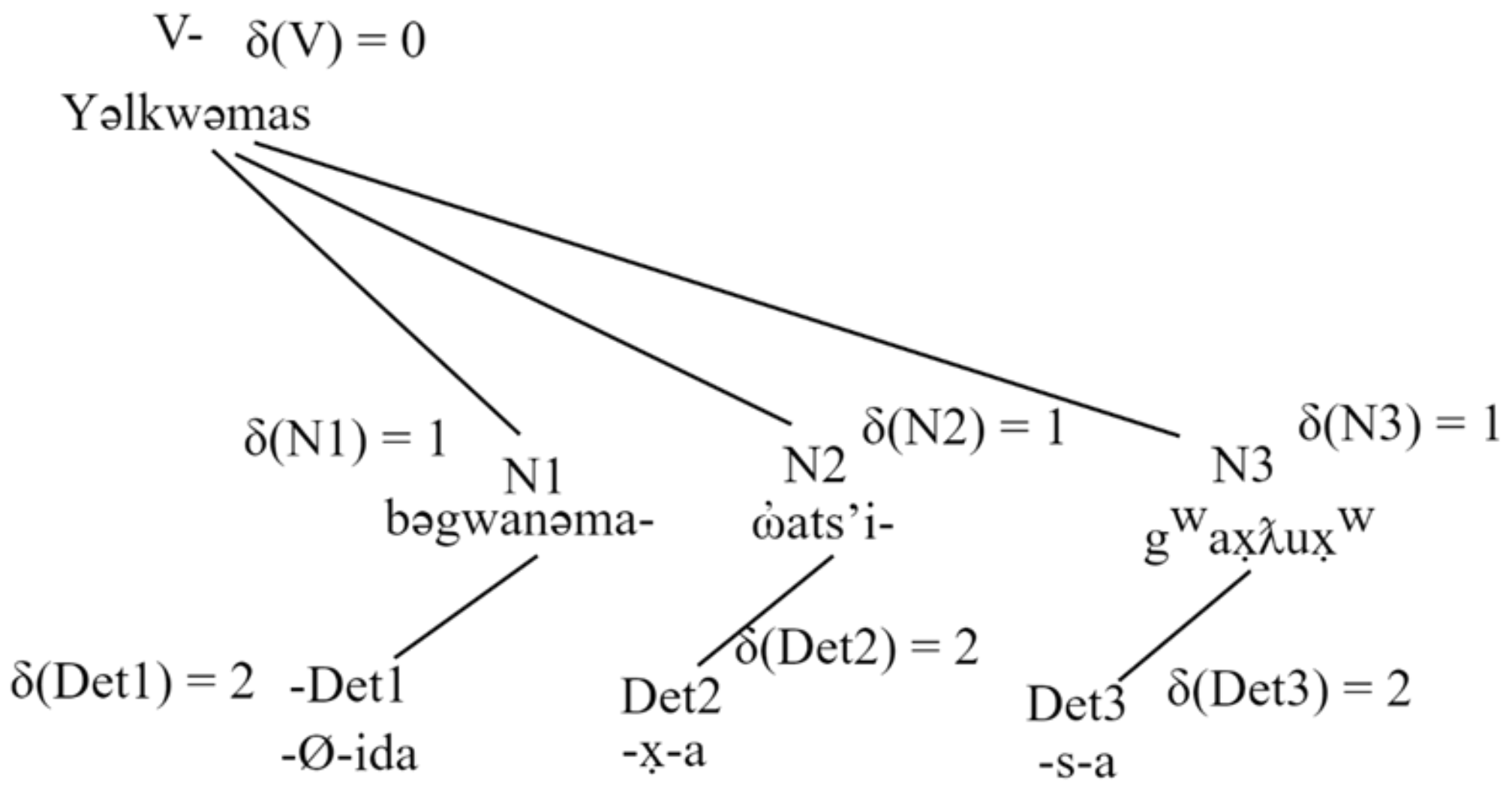

Fig 5. The DG graph of (7)

In Figure 5, we have shown the dependency hierarchy along with the linear relations of the words/morphemes. We would also adopt this convention for the rest of the paper for ease in understanding. The dependency graph above makes it evident on the one hand that the nominative case marker (*-Ø-*) and the demonstrative *(-ida*) qualify the N *bəg*$^{w}$*anəma-* but not the verb *Yəlk*$^{w}$*əmas-* they are attached to and on the other that the interpolated subject disrupting the string continuity of the verb and the object as part of the verb phrase in a constituency-based analysis does not trigger the same problem in dependency relations which are intrinsically (linear) order-free, as also pointed out above.

Now we may turn to the CG representation of the same sentence in (7c). It is vital to emphasize here that in our CG analysis of sentences the derivations may involve functors whose arguments are not (always) adjacent to the functors due to discontinuity. In such cases, the functor and the argument as part of a discontinuous expression at hand are shown to work in the usual manner of functor-argument applications (regardless of the presence or absence of discontinuous expressions/constituents). The only proviso here is that the functor and the argument as part of a discontinuous expression can be interpreted as a combined lexical form when produced as an output of 'wrapping[2]' in order that the output yields a category that can take an adjacent argument (Morrill 1995, 197-198; see also Steedman 2014, 682-683). In this paper, we have not adopted type-raising as an operation for discontinuous structures because we believe this introduces an arbitrary amount of complexity into the analysis of specific word categories, for the number of categories would arbitrarily increase depending on different linguistic constructions. Hence, we have allowed for wrapping over more than one constituent for discontinuous structures because any kind of wrapping over more than one constituent can be accomplished by means of the *iterative* application of the operation of wrapping[3]. For example, in (7c) the verb *Yəlkʷəmas-* and the object NP *x̣-a ὠats'i-* can form a sort of a wrapping functor which can then take the subject NP as an argument: (*Yəlkʷəmas-* + *x̣-a* +*ὠats'i-*)--(S/N). Although this is not explicitly shown in the CG derivation below, this interpretation may be taken to be presupposed for all CG derivations in this paper.

(7c) The CG derivation:

[2]Wrapping rules usually infix, by way of a sort of swapping, a discontinuous string element in a place where another element was initially located (see for details, Steedman 1985).

[3] For instance, in the case of the verb phrase *shrug Jenny angrily off*, the sequence of categories is V-NP-AdvP-Particle (P). If we want to apply wrapping to the categories V and P, the wrapping operation would be over two constituents, i.e. NP and AdvP. This can be accomplished by applying wrapping iteratively in the following manner: Step 1: (V+AdvP); Step 2: ((V+AdvP)+P). This output category ((V+AdvP)+P) as a functor can then take the NP as an argument.

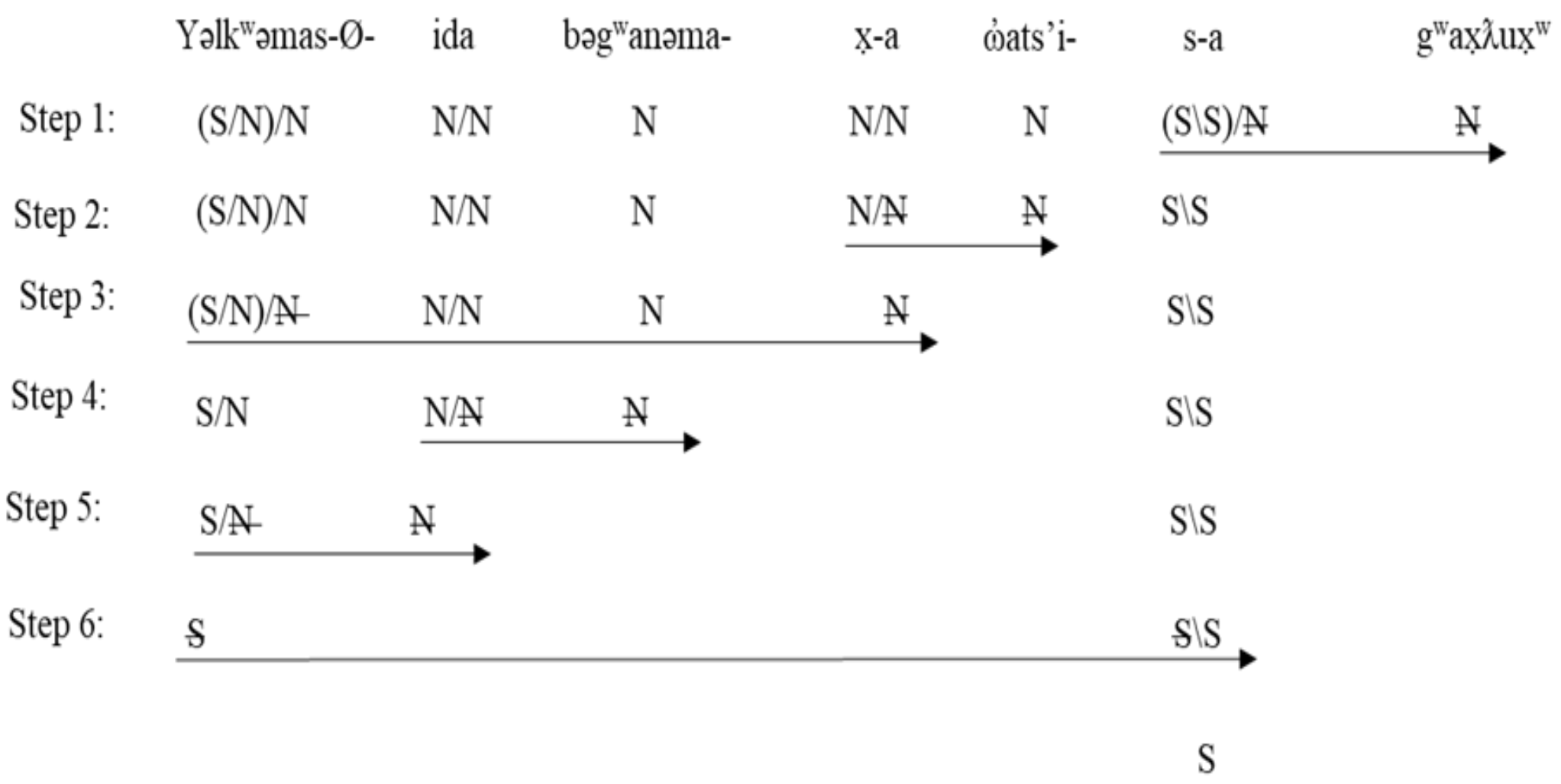

Fig 6. The CG derivation of (7)

Since PSG prohibits crossing branches, the case marker and the determiner of the noun are analysed as separate constituents/parts along with the root verb as seen in Figure 4. However, as seen in Figure 5, the demonstrative (which is attached as an affix to the verb) is actually dependent on the noun, not on the attached root. Even in the corresponding CG analysis, the affix and the noun are cancelled out (not the verb and the affix), as can be seen in Figure 6, thereby accounting for the fact that the determiner is dependent on the noun, not on the verb to which it is attached. This clearly shows the advantages of using DG and CG for accounting for structures of the languages showing discontinuity.

Therefore, this suggests that these two types of formalisms (constituency-based and non-constituency based) possibly form the extreme ends on the scale of formalization of the level of (dis)continuity, reflecting whether linguistic expressions are analysed either in terms of only constituents or in terms of non-constituent relationships. However, in the real world a speaker of a language with a predominantly continuous system can also learn, comprehend, speak a language with a non-continuous system. Though linguists advance each of these grammatical formalisms on distinct grounds having varying theoretical motivations, it is more likely that for a speaker of any language there exists just one representation in their cognitive system which is equipped to deal with the features of both kinds of systems. The present work aims to approach the problem of discontinuity by integrating the grammatical formalisms which address continuity with those addressing discontinuity. This raises a natural question of why the focus is on only these three formalisms. Part of the reason is purely practical- to make the job more manageable by narrowing down the domain of inquiry to these three formalisms. Both DG and CG handle discontinuity but in different ways. DG is based on dependency relations and CG is based on functor-argument relations. However, contrary to this widely held assumption, both DG and CG are not after all distinct from one another because dependency relations and functor-argument relations can be mutually united too. Along similar lines of thinking, the idea of ‘dependency constituency’ has been formulated by Barry and Pickering (1990, 1993), although the present conception differs radically from theirs, as will become clearer in section 6. Most importantly, what makes this approach different from the earlier ones is that it attempts to face up to the problem of discontinuity without introducing any extra assumptions/rules or even constraints. The very flexibility that PSG can have in allowing for both normal trees and tangled trees (by relaxing the ‘no-crossing’ constraint) is nothing other than the flexibility DG or CG inherently permits. DG or CG is inherently neutral with respect to line crossing or no-line crossing. Hence the desired flexibility in PSG for continuous and discontinuous structures is an expression or instantiation of the principles of DG/CG itself. Thus, the apparent tension between the three grammar formalisms can perhaps be neutralized by this way of working towards mutually unifying *the most basic principles* of the three formalisms. Before articulating the core formulation of this paper, we shall discuss different existing approaches to discontinuity proposed so far in the literature.

**5. Questioning the Universal Validity of a Rigid Constituent: Previous Approaches to Discontinuity**

Traditional generative grammar greatly emphasizes the interdependence of grammatical relations, thematic roles and phrase structure. For example, once Agents are encoded as subjects, they are encoded as subjects in all transformations. This is ensured by Chomsky's (1981) Theta Criterion (ensuring a one-to-one mapping between a theta role and an argument) and Baker's (1988) Uniformity of Theta-Assignment Hypothesis (ensuring a a one-to-one mapping between thematic relations and structural positions at pre-transformational stages). Thus, a unique mapping between theta roles and grammatical functions has been considered essential. The idea has been that all sentences are to be analysed in terms of continuous constituents/phrases. This widely accepted notion of phrase structure constituency is, to a large extent, motivated by and so aligns with the familiar understanding of NP, VP, AP, PP as projections of their heads N, V, A and P respectively, and of the nominative-accusative case marking languages such as English. This is not to deny that PSG analyses of ergative languages are available. A sentence is thereby the projection of Inflection (Tense) (Chomsky 1995, Newmeyer 2001). As already stressed in earlier sections, the grammar formalism primarily based on constituency is Phrase Structure Grammar (PSG). All word order variations are to be encoded in the grammar rules directly. There are two important general structural relations in PSG: Dominance and Linearity. However, this formalism helps us in explaining word order governing rules in languages such as English with somewhat inflexible word order (also known as configurational languages). They allow for continuous constituents (Gazdar 1983). Can one universally apply this kind of understanding of a constituent? The last few decades have seen several linguists arguing for alternative approaches towards solving the problem of discontinuity. Below we discuss various approaches proposed thus far to account for discontinuity in natural languages (Austin and Bresnan 1996, Legate 2003, see also, Müller 2004).

### 5.1 Austin and Bresnan on Warlpiri's phrase structure and Lexical Functional Grammar

An alternative approach towards analysing Warlpiri sentences using a hierarchical structure proposed by Austin and Bresnan (1996) made two important claims: Warlpiri phrase structure is flat and is characterised by free base-generation of elements in any order within the clause (although constituent structure did not matter much (see also, Bresnan 2001, 110)). Further developments led to the formulation of Lexical Functional Grammar (Kaplan and Bresnan 1982). We shall show below how LFG handles discontinuity through one illustrative example.

#### 5.1.1 Lexical Functional Grammar (LFG)

The basic assumption of the derivational theory of grammar is that there exists a mapping between the predicate argument structure and the phrase structure configuration. The motivation for this assumption has been derived from the relationship exhibited by active and passive sentences in English. For example, *The cat killed the rat* (active) and *The rat was killed by the cat* (passive). The NP *the cat* preceding the verb is the 'agent' in the active sentence and 'theme' in the passive sentence. It was thus postulated that movement of the NP following the verb to the position in front of the verb would result in the passive form; accordingly, two different sets of rules exist for the two forms (active and passive). This assumption certainly has its limitations. In non-configurational languages, the grammatical functions (subject, object etc.) are present with morphological case marking, not as distinct

NPs. In such situations, passivization does not involve movement of a constituent (NP), but it involves a change in morphological case.

Arguing against this assumption that two different sets of rules exist for the active and passive forms, Kaplan and Bresnan (1982) put forth the argument that if one knows that only in the active sentence the subject is the agent and the object is the patient, based on the lexical entry of the verb, the information for both the forms can be represented (instead of two separate sets of rules). This is because, by implication, in the passive sentence, the subject becomes the patient and the *by*-object (Oblique) is the agent. The theory of LFG was thus formulated in response to the limitations in the derivational theory, by placing an emphasis on the lexical assignments of grammatical functions guided by the universal principles in the predicate argument structure (Bresnan 1982, Dalrymple 2002). Hence, this theory rejects the traditional transformational grammar's one-to-one correspondence between the predicate argument structure and the constituent structure and provides an alternative model of syntax for accounting for discontinuity in natural language. In order to show how discontinuity is handled in LFG, we discuss below LFG's account of long-distance dependencies (for example, topicalisation) from English (Kaplan and Zaenan 1989).

Consider the multi-complement sentence *Mary, John claimed that Bill said that Henry telephoned.* In this multi-complement sentence, *Mary* is originally understood as the complement of the verb *telephoned.* Its c-structure is depicted in Figure 7:

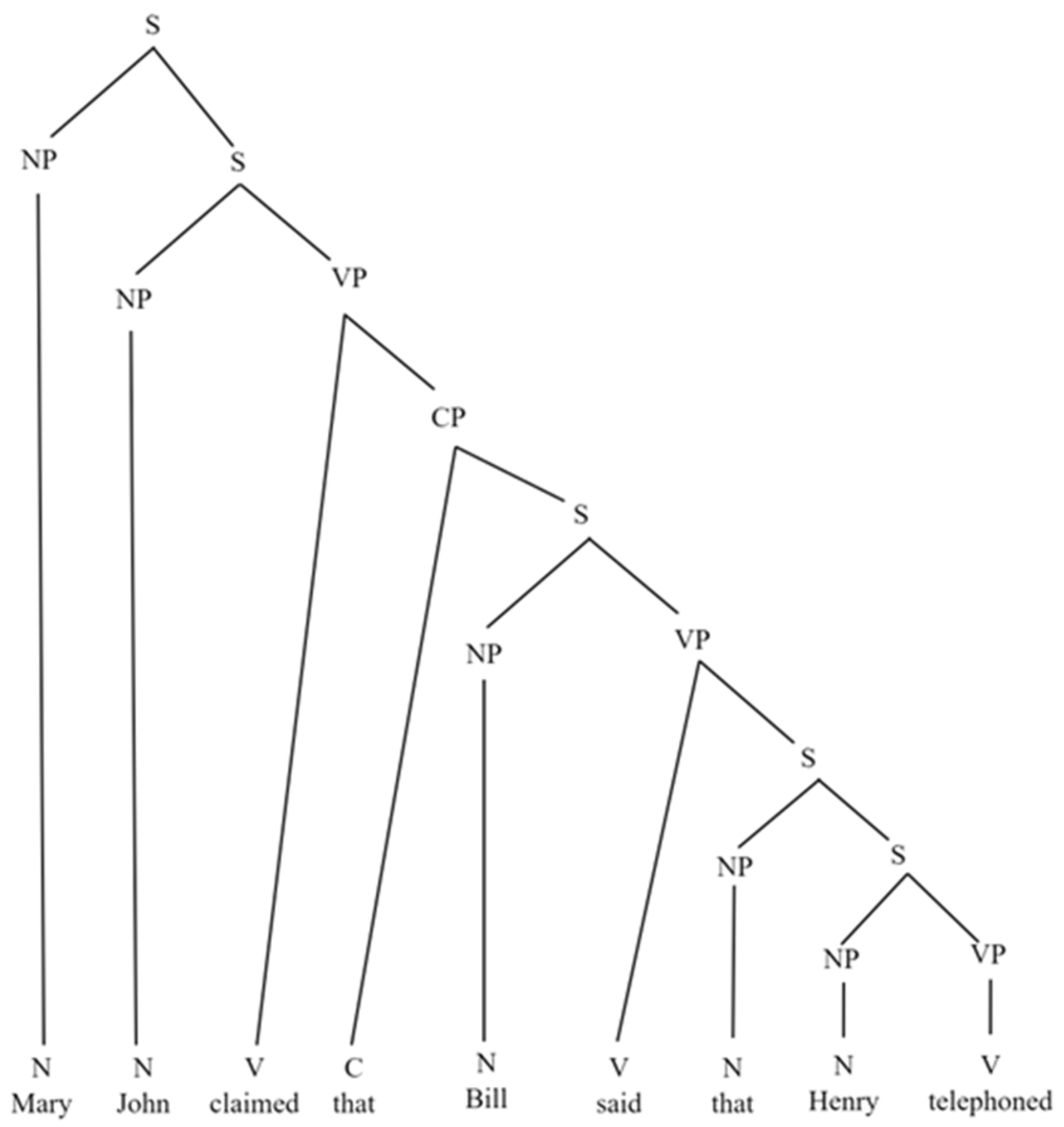

Fig 7. The c-structure of *Mary, John claimed that Bill said that Henry telephoned*

An important observation to be noted is that the c-structure does not depict an empty NP node as the complement of the final VP, which is otherwise present in a PSG tree. The fact that the object of the most deeply embedded clause has undergone topicalisation is shown in the f-structure, as seen in Figure 8:

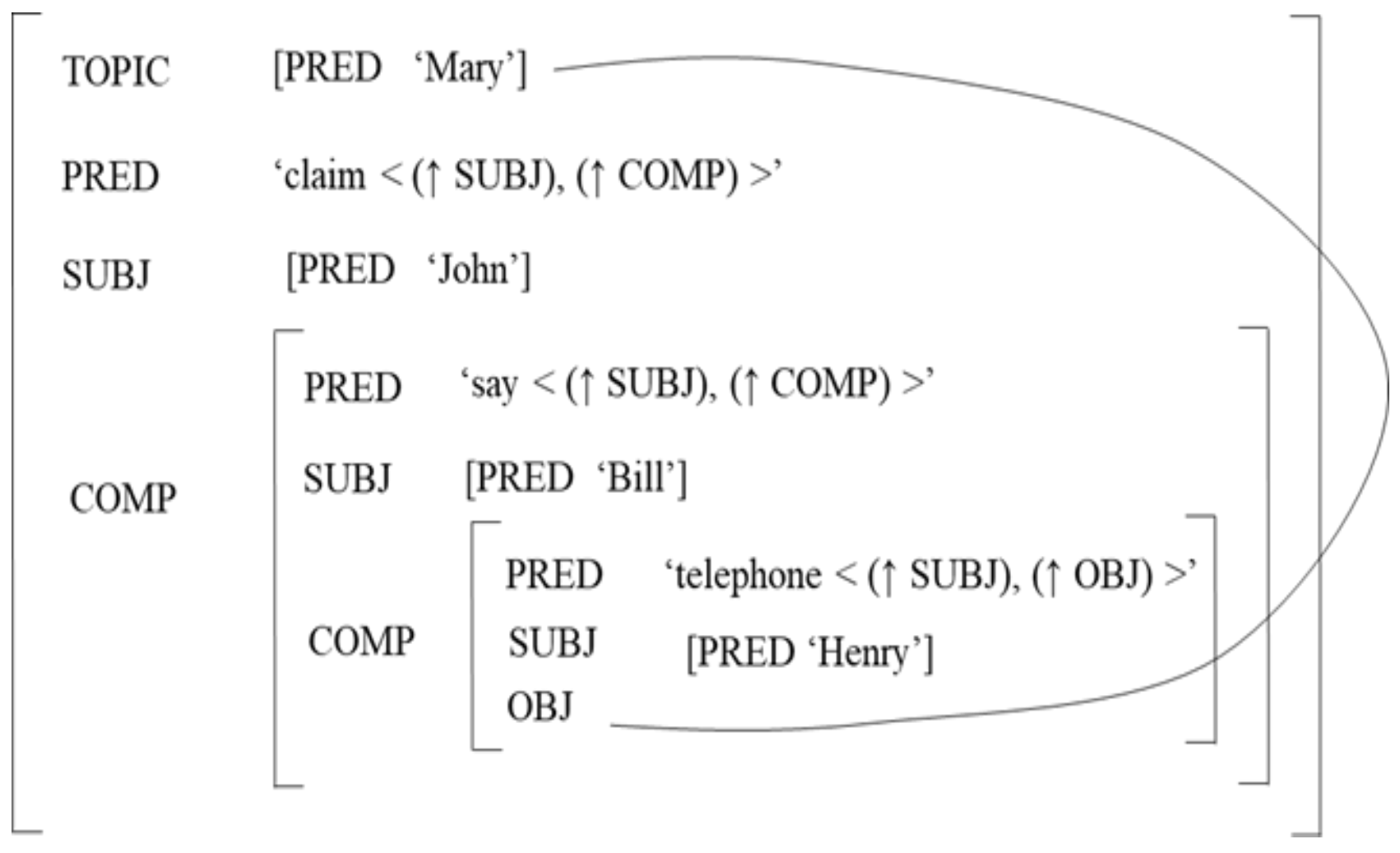

Fig 8. The f-structure of the above sentence

*Functional Control:* The AVM has four attributes: PRED, SUBJ, TOPIC, COMP(lement). COMP is the grammatical function of clausal complements. Its value consists of the attributes SUBJ, PRED, OBJ. The link in the f-structure depicts the relationship between the topic and the object of the most deeply embedded complement. Thus, a single part of an f-structure plays two or more roles in the f-structure, and functional control can occur over several layers of embedding.

### 5.2 Phenogrammatical structure

Another alternative approach towards discontinuity was proposed by Dowty (1996) who makes two important assumptions. First, he proposes a 'minimalist theory of syntax' to describe various discontinuous syntactic phenomena by taking linear structure as the norm rather than hierarchical structure, that is, 'a clause or a group of words is *only* a string'. Second, some words and constituents are more tightly bound (attached) to adjacent words than others. The linear structures/representations of expressions are treated as unordered lists, as seen in Figure 9.

(8) *A woman who John knew arrived*.

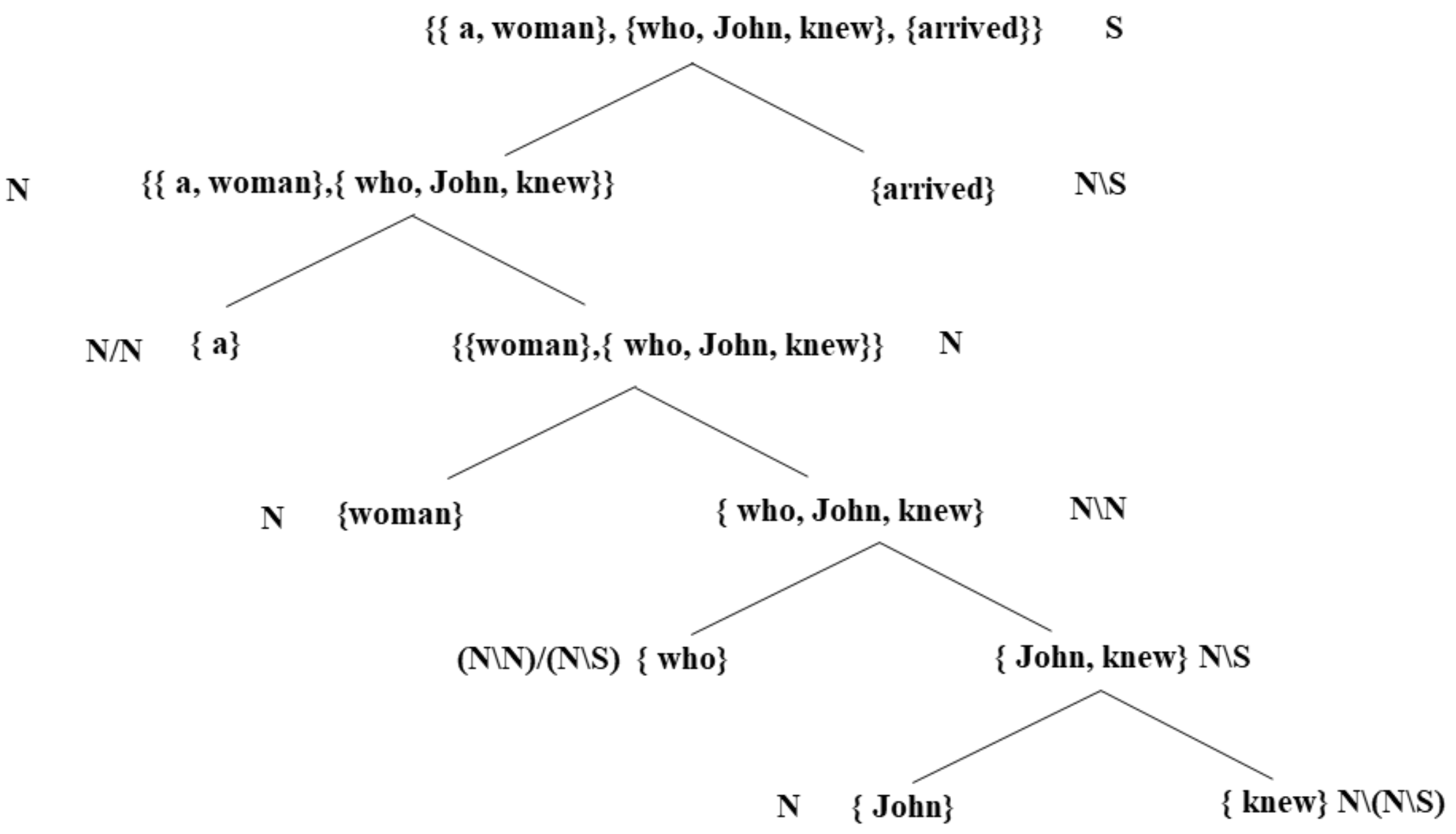

Fig 9. Phenogrammatical structure of *A woman who John knew arrived*

These unordered lists represented in the form of a set union are broken to form linear sequences that comply with the word order rules (precedence or succession relations) in the language. Dowty calls this level the ‘phenogrammatical structure’. The manner in which the whole structure is built up from the component expressions is termed the ‘tectogrammatical’ structure. In the phenogrammatical structure, the role of word order in expressing syntactic organisation, along with the role of inflectional morphology, is articulated. This kind of representation is built on the rules of CG and a new system of non-concatenative enumeration of lexical items. Along similar lines, Donohue and Sag (1999) have adopted Reape’s (1996) ‘sequence union operation’ or ‘shuffle’ for the analysis of discontinuous constituents within the framework of Head-driven Phrase Structure Grammar. The sequence union of two lists $l_1$ = <a,b> and $l_2$ = <c,d> is the list $l_3$ iff each of the elements in $l_1$ and $l_2$ is present in $l_3$ and the original order of the elements in $l_1$ and $l_2$ is preserved. For example, the sequence union of $l_1$ and $l_2$ is any of the following lists/sequences: <a,b,c,d>, <a,c,d,b>, <a,c,b,d>, <c,d,a,b>, <c,a,d,b>, <c,a,b,d> but not <b,a,c,d>, <a,b,d,c> etc. This allows discontinuous elements to intervene in the linear order of a constituent, thus accounting for discontinuity.

**5.3 Tangled trees**

Another approach towards discontinuity and crisscrossing dependencies has been discussed by McCawley (1982, 1987), Iwakura (1988), and also by Blevins (1990). McCawley argues for allowing transformations which alter word order without causing any change to the constituent structure to yield *discontinuously projected constituent* structures. The idea of constituency is retained, not completely eliminated, but with relaxations on two conditions/constraints (which are disallowed by the traditional formalism of generative grammar): (i) no-crossing constraint (ii) single mother condition (the condition that each node will have a single mother). He thus relaxes these standard axioms in order that both multidominant trees and trees with crossing nodes can be permitted. This results in 'tangled' trees in which the problem of discontinuity is accounted for and also the advantage of a 2D representation is preserved. Importantly, he redefines the notion of c-command. The usual definition of c-command is: a node X c-commands a node Y *iff* the first branching node above X dominates Y. McCawley argues that a node X will c-command not only node Y which is dominated by X’s mother but also nodes dominated by the modifiers of the bounding node (X’s mother). Fig. 10 and Fig. 11 are the phrase structure and tangled tree diagrams, respectively, of the same sentence: *Tom may be, and everyone is sure Mary is, a genius*.

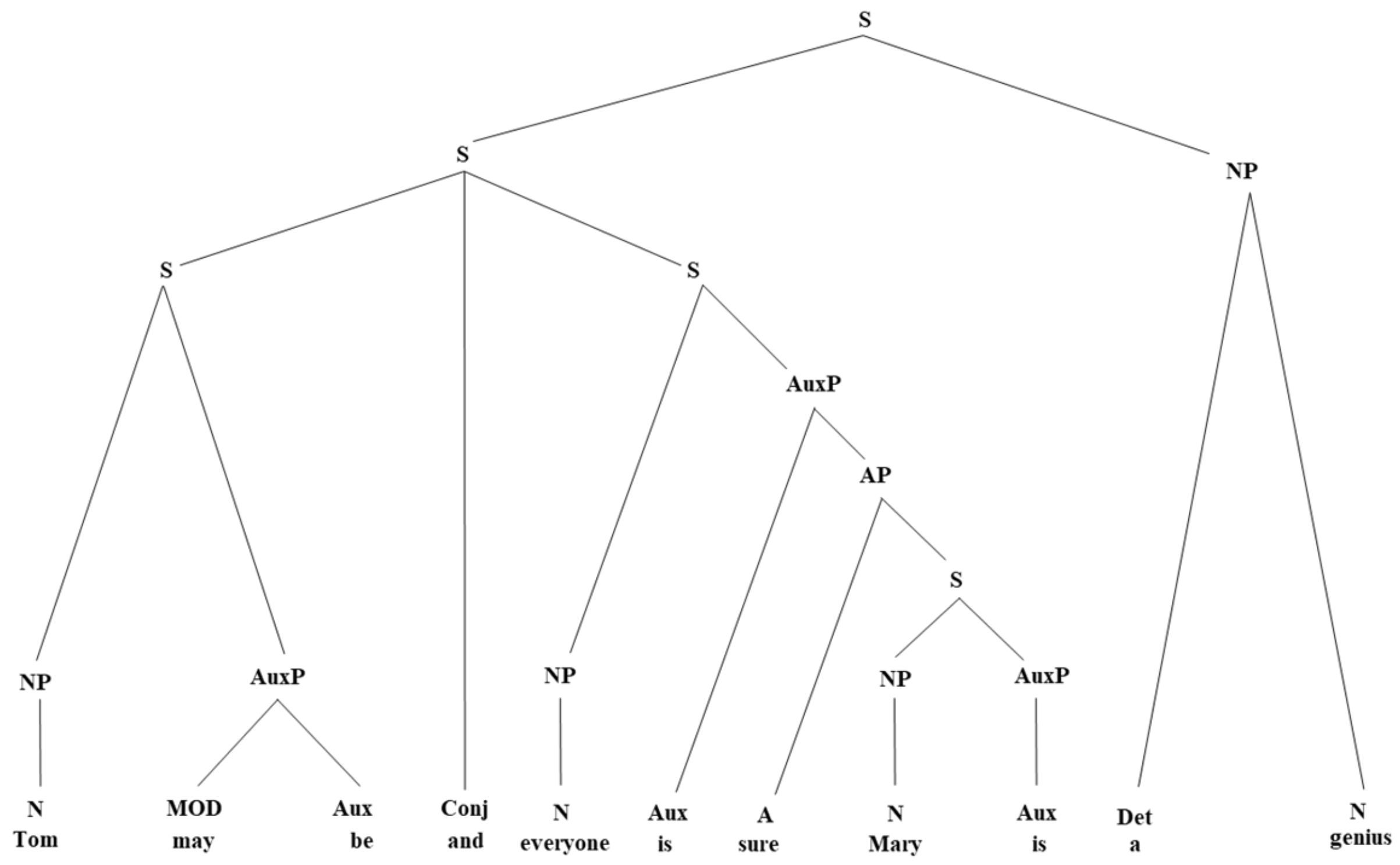

Fig 10. A PSG diagram of *Tom may be, and everyone is sure Mary is, a genius*

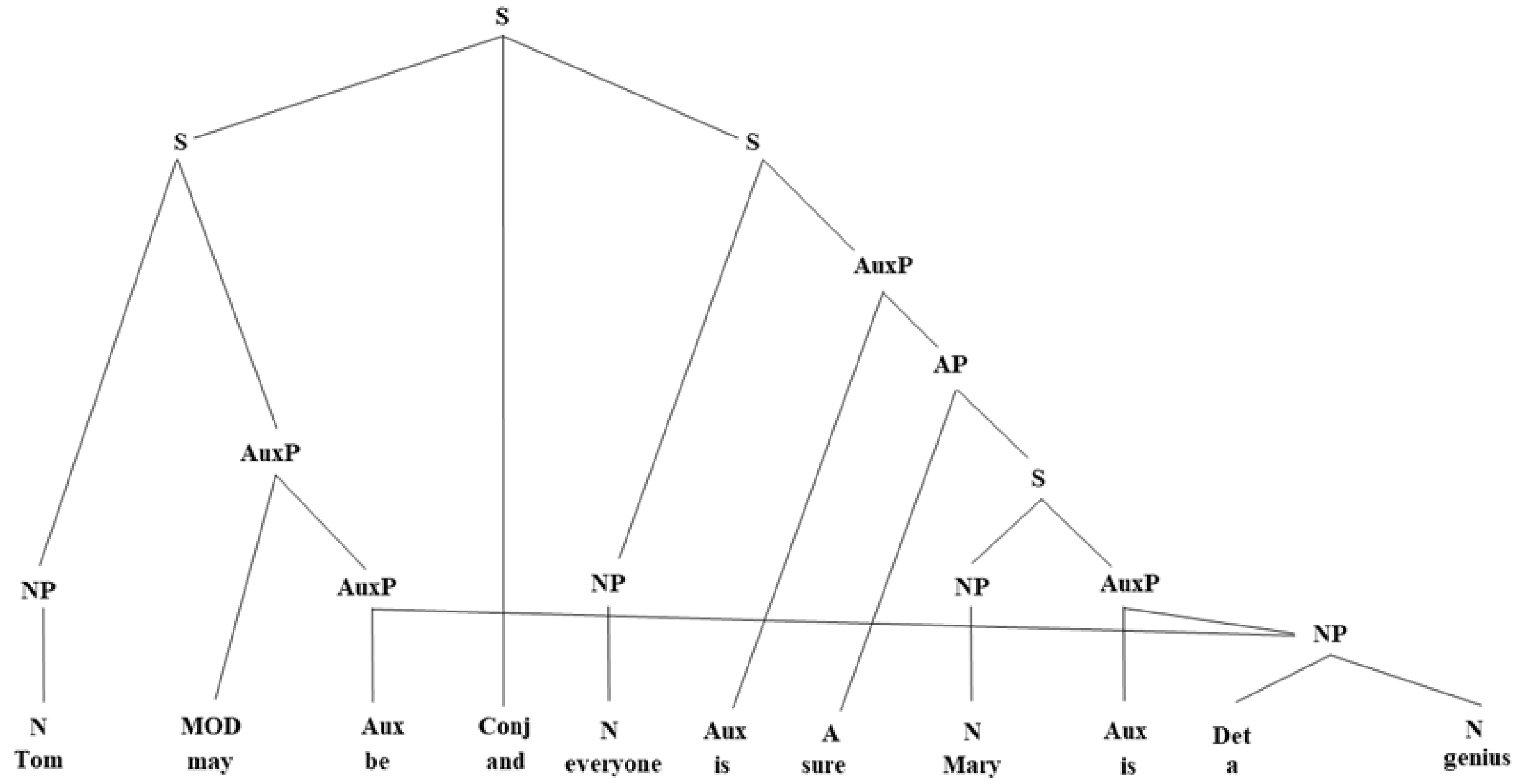

Fig 11. A tangled tree diagram of the same sentence (as per McCawley's proposal)

As clearly seen, the NP *a genius* is analysed as a constituent in the tangled tree as well. The difference is that in Fig. 10 *be* and *is* do not c-command *a genius*, whereas in Fig. 11 they c-command *a genius*. Thus, by relaxing the no-crossing constraint and the single mother condition assumed in PSG, tangled trees try to solve the problem of discontinuity and crisscrossing dependencies.

**5.4 The approach of Parallel Merge**

A similar kind of solution was proposed by Citko (2011) in cases where one element is shared by two nodes, though similar, but in a slightly different way from McCawley's approach of tangled trees. In this approach called *Parallel Merge,* the Single Root Condition is relaxed, multidominance is derived without making any extra assumptions. She discusses ways to linearize multidominant structures (which are typically considered to be non-existent in the mainstream generative syntactic theory). It is explained to be implemented in two steps as seen in Figure 12.

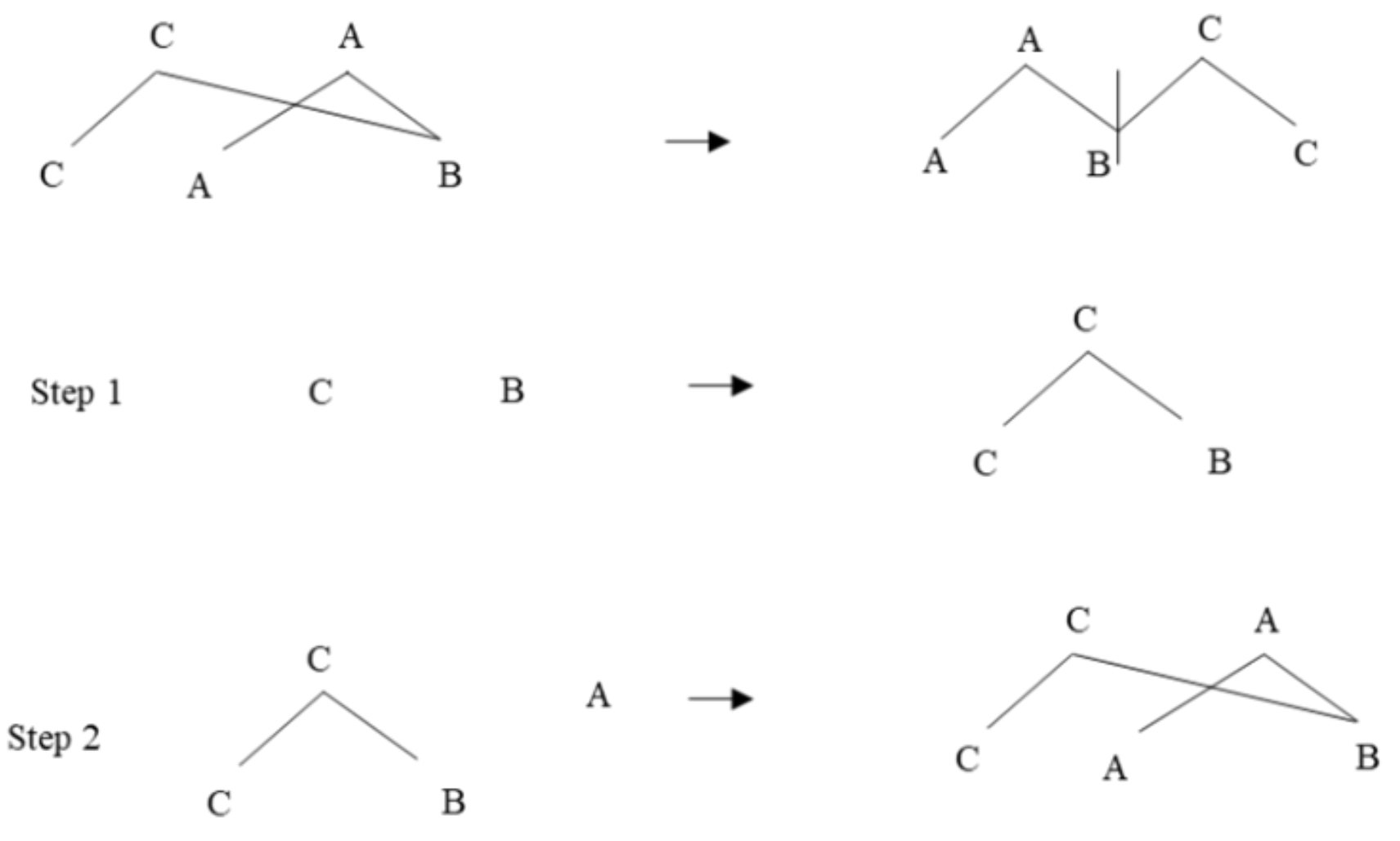

Fig 12. Illustration of Citko's Parallel Merge approach

In the first step, node C combines with node B and in the second step node A combines with the output of step one. Node B is considered to be the PIVOT between the nodes A and C. According to Citko, the symmetry inherent in the above structure can be understood as a property of the c-command relation, that is, the pivot B stands in a symmetric c-command relationship with respect to the other nodes A and C (sisterhood relationship). There are no restrictions on the pivot since any element can be a pivot.

Citko also advocates the need for tangled trees in examples where the same NP is sort of *shared* by two predicates in the conjuncts as seen in Figure 13 for the sentence *My friend likes, and my sister hates, this kind of movies*.

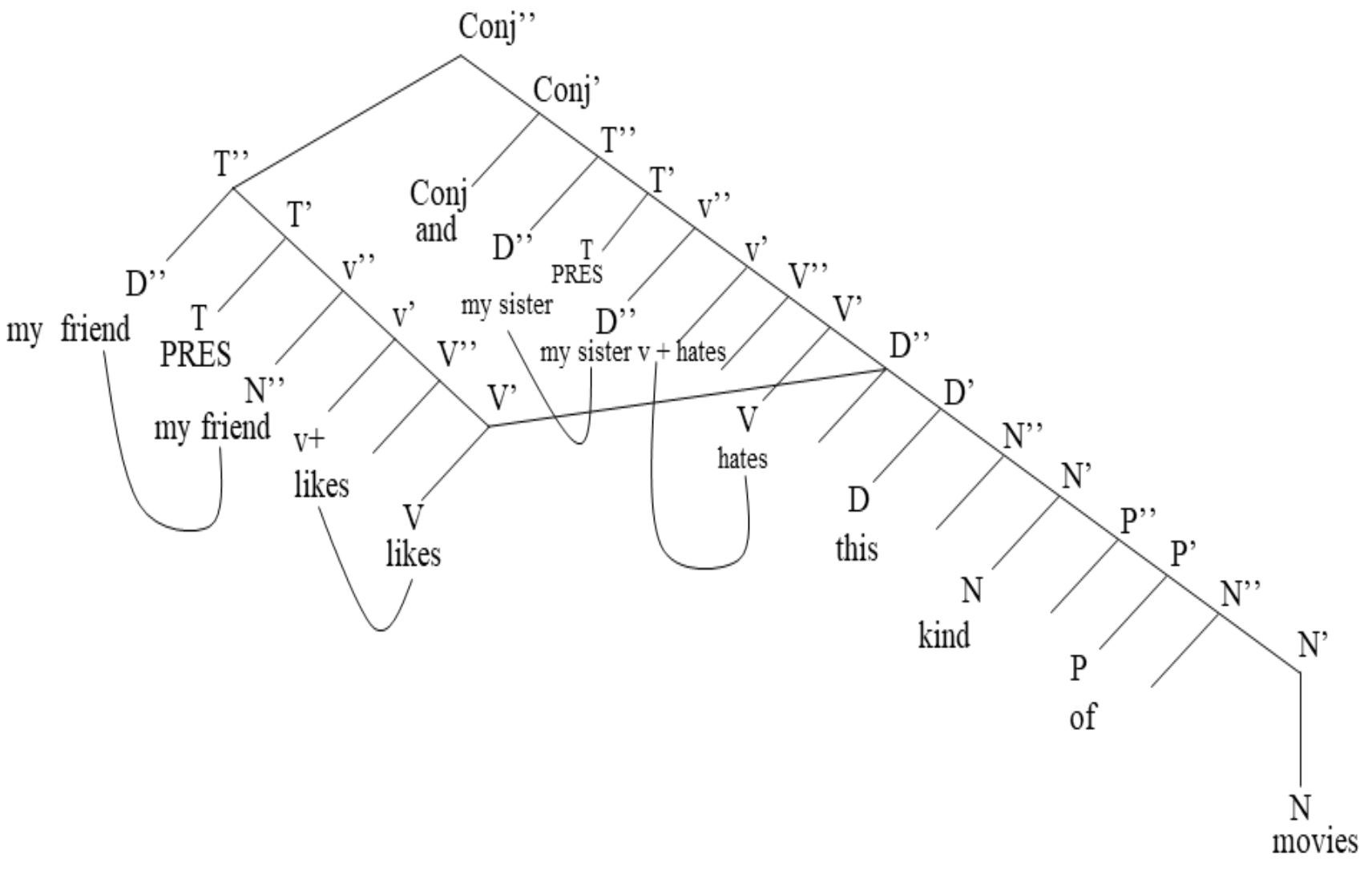

Fig 13. A tangled tree diagram of the sentence: *My friend likes, and my sister hates, this kind of movies*.

Significantly, this approach slightly differs from McCawley's approach. As per Citko's analysis, the symmetry involved in this structure is quite obvious: The pivot *'this kind of movies'* stands in a symmetric relationship with other elements inside the two conjuncts and is c-commanded by the material inside both the conjuncts. (To illustrate Citko's point, the diagram above is sketched out in terms of X-bar phrase structure as assumed by Citko, and sentences are represented in highly articulated tree diagrams in X-bar phrase structure.)

Citko also offers novel analyses to account for a symmetric multidominant structure in a free relative construction. For example, in the sentence *I read what(ever) Bill wrote,* the *wh-*pronoun *what(ever)* is a pivot between two complementizer phrase (CP) chunks: [ $_{CP}$ *I read whatever*] and [ $_{CP}$ *whatever Bill wrote* $t_i$]. The shared position is the specifier of the relative CP ('whatever Bill wrote') and the complement of the matrix verb ('read'), and thus this gives rise to a symmetric structure as seen in Figure 14.

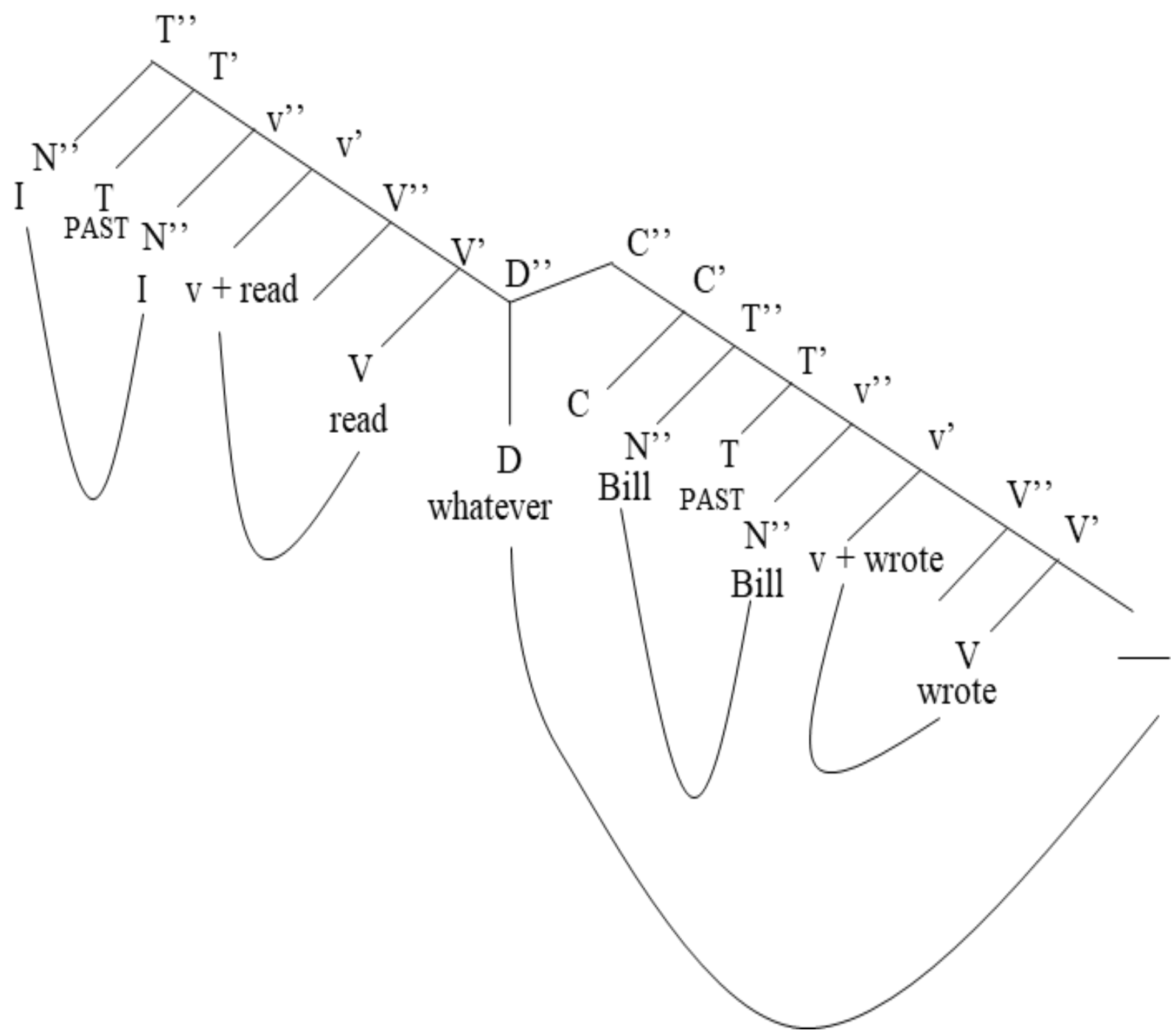

Fig 14. An analysis of the sentence *I read whatever Bill wrote*

**5.5 Dependency constituents**

Barry and Pickering (1990, 1993, 864-5) have attempted to show the constituent relations in the phrase structures with dependency relations. They call such a flexible constituent a ‘dependency constituent’. A dependency constituent consists of all words linked by dependencies. If its words form a substring of the sentence, it is a continuous dependency constituent. Consider the sentence:

(9) *John thinks Bill laughed*.

The labelled bracketing for this sentence is: [$_{S}$ [$_{NP}$ *John*] [$_{VP}$ [$_{V}$ *thinks*] [$_{CP}$ [$_{C}$ φ] [$_{S}$ [$_{NP}$ *Bill*] [$_{VP}$ *laughed*]]]]]. Accordingly, *thinks Bill laughed* and *Bill laughed* are also constituents (VP and S respectively) apart from the individual words and the entire sentence as per the conventional phrase structure. The head-dependent relations (as per DG) in this sentence are as seen in Figure 15:

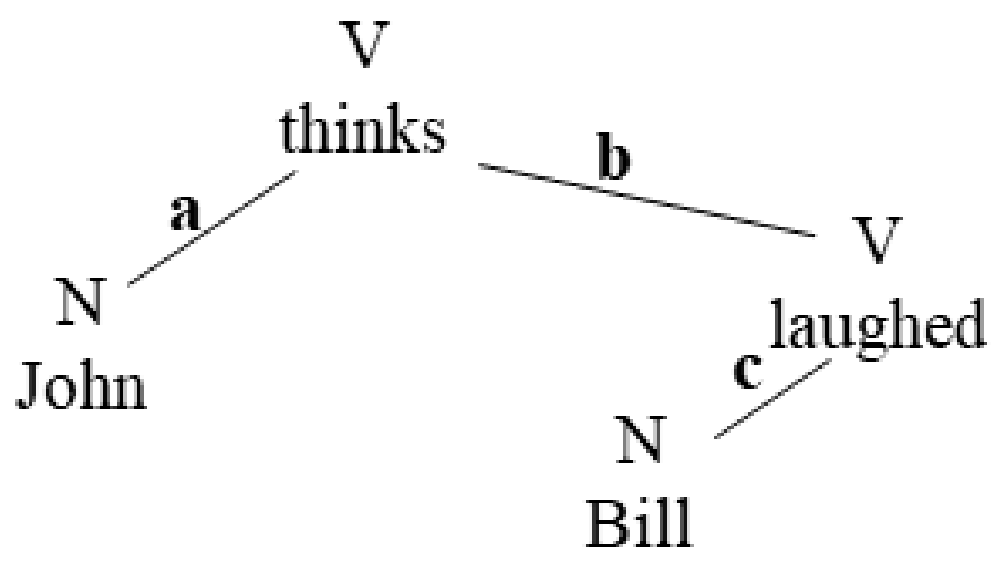

Fig 15. A dependency graph of *John thinks Bill laughed*

But according to Barry and Pickering (1990, 1993), the dependency constituents for this sentence would be the following.

i. *John thinks* (as indicated by ‘a’)

ii. *Bill laughed* (as indicated by ‘c’)

iii. *thinks Bill laughed*

iv. The individual words – *John, thinks, Bill, laughed*

v. The entire sentence *John thinks Bill laughed*

These are the continuous dependency constituents since they all form substrings of the sentences without any break. Also, *thinks* and *laughed* form a dependency constituent, but discontinuous. *John thinks* and *thinks Bill laughed* are overlapping dependency constituents with *think* as the overlapping string. Therefore, it moves a step ahead of the traditional understanding of a constituent in accounting for the dependencies which are not usually considered to be a constituent, thereby allowing for flexibility.

At this juncture, it is also noteworthy that Hays (1964: 513) has discussed ways of encoding a constituent in phrase structure theory in terms of dependency theory. A dependency rule V(N1*N2, Adv) corresponds to an expression such as *Athletes played badminton boldly*. Thus, subtrees can be isolated from DG trees corresponding to constituents in PSG graphs. Gaifman (1965: 316-325) has notably demonstrated that there exist formal correspondences between DG structures and PSG structures, in that ramifications of PSG rules called ‘parenthetical expressions’ help establish equivalences of some sort between dependency trees in DG (also called a 'd-system') and a unique constituent in a PSG tree (called a ‘p-system’) and vice-versa. Analogous explorations have also been made in terms of 'structure-free[4]' representations of strings between PSG and DG systems in Robinson (1970: 260-263). However, these derivations are limited to showing only the correspondences between PSG and DG (see Müller 2020). In the present paper, we go beyond this and attempt to unify CG functor-argument relations with DG head-dependent relations so that they can be smoothly integrated with constituency rules in PSG. The next section shows elaborate steps of arriving at the unified system of representation.

## 6. The Unified System of Representation

The idea is to integrate the representations of structure in the three formalisms into a single system of representation that can handle both continuity and discontinuity. Grammars which allow for continuous constituents require many rules to be written and grammars which allow for discontinuity are analysed in terms of non-constituent relations. However, we have the same cognitive system, for example, for a speaker of English (a language with a continuous system) and also for a speaker of Warlpiri or Navajo (languages which allow for discontinuous constituents). Ultimately, it is the same mind/brain which represents the two different grammars. It has neither *only* the capacity to store an unlimited number of rules for continuous structures nor *only* the capacity to analyse clauses in terms of lexical items and their dependencies and/or functor-argument relations. The cognitive system cannot adopt multiple representations for both continuous and discontinuous structures within and across languages. The problem persists even within a single language, since a language that usually has continuous structures can have discontinuous structures as well (parenthetical structures/expressions (section 1), topicalization, sharing, etc. in English, for example). Hence it is plausible that there exists one system of representation which incorporates both types of grammars. The aim is to unify the general constraints and principles of continuous and discontinuous grammar formalisms. This may have implications for the possible cognitive organisation of the continuous and discontinuous systems. We would not pursue this line of inquiry further as this is beyond the scope of the paper.

[4]Structure-free representations of structure are intuitively those that appeal to the uniqueness of the replacing string with respect to a grammatical category to be distinguished from another. For instance, Subject--> NP and Object--> NP are *not* structure-free representations.

Three examples from German, Croatian, and Kalkatungu have been taken for the illustrative purpose of making the derivations that can show the desired equivalence perspicuous. Just to provide a brief context of these languages, we may note that free word order in German is quite well-known with both OS and SO sequences permitted (see Bader and Häussler 2010), while word order flexibility in Croatian and Kalkatungu is a bit different. On the one hand, Croatian, a language of the southern branch of the Slavic family, has an unmarked order of SVO with the contextually determined flexible ordering of the predicate and its arguments, which is often governed by the theme (less informative)-rheme (more informative) sequence and its reversal in certain contexts (Siewierska and Uhlířová 1998). Also, clitic placement in Croatian is usually in the second position in accordance with Wackernagel's law. On the other hand, Kalkatungu, an Australian aboriginal language of western Queensland, has a very flexible system of word order—it is not just the predicate and its arguments that can be arranged in any order, all words that make up an argument can appear in any order (Blake 1983).

Examples (10-12) are illustrations of tree diagrams for discontinuous expressions from German, Croatian, and Kalkatungu (respectively) depicting the CG derivation in PSG trees and the dependency functions in terms of CG formulae. To establish the equivalence relations, the converse derivations have also been worked out and, finally, we arrive at a unified representation from DG to CG to PSG at the end of the series of derivations.

Before we proceed to show the desired way of uniting the representations, we propose a principle that would help unite the DG and CG representations. We need this principle precisely because CG derivations would piggyback on PSG constituents in our analyses but the CG relations defined on these constituents have to be mapped onto the dependency relations[5]. The principle that we propose below is exactly the one that will serve this role. We call this the *Correspondence Principle*. This would be used for the DG → CG and CG → DG derivations illustrated below.

*The Correspondence Principle:*

Below is *The Correspondence Principle* (Anonymous 2022, Anonymous 2023, Anonymous 2024) proposed to unify the direct head-dependent relation and the functor-argument relation between any two given words A and B:

$$A(B^*) \vee A(^*B) \equiv A|B$$

Here, A(B*) indicates B is dependent on A and B is to the left of A and A(*B) indicates B is dependent on A and B is to the right of A as per the linear order of the sentence. Here, '∨' is the logical disjunction, '≡' is a special equivalence sign and A|B indicates that either A or B can be the functor in categorial relations, with '|' indicating the neutral direction of the functor. This implies that the other element will be the argument. The logical relation is that of an *implication*, but not of *an entailment*, because when one element (A or B) is the functor, nothing is said about the other element. In cases where there is a direct dependency relation between the functor and the argument, A and B on the Left-Hand Side (LHS) and Right-Hand Side (RHS) turn out to be the same. However, this is not the case always. In exceptional cases, only either A or B tends to be the same on LHS and RHS, and the other category can vary across sides. If, for example, we suppose that A is the same on both sides, the exact value of B may differ on the LHS and the RHS (that is, B can take a word X, for example, on the LHS, while it takes a word Y, for example, on the RHS). Thus, it is not always the case that the head word is the functor, and the dependent is the argument.

[5]This principle is supposed to filter out all relations between words that do not conform to functor-argument relations or dependency relations. Consequently, the relations that will be filtered out will not also form constituency relations in PSG. For example, *I like the classroom.* The relation between words such as *I* and *the* or *I* and *classroom* will be filtered out after the *Correspondence principle* is applied, precisely because these words do not enter into a functor-argument or (direct) dependency relation. Therefore, this principle itself imposes a constraint on the conceivable relations between words in sentences/clauses.

A simple illustration of the principle with the help of the Warlpiri example in (1) can be provided. At the level of the VP, the verb *panti-rni* bearing the category (S/N)/N cancels out the category of the nominal expression *wawirri* (that is, N), while *wawirri* is dependent on *panti-rni*. Thus, we have that (S/N)/~~N~~ ~~N~~ ≡ *panti-rni* (* *wawirri*). The tangling in the PSG tree that may appear at this point will be just the manifestation (or epiphenomenon) of this correspondence. Then at the level of the AuxP, the auxiliary *ka* bearing the category (S/N)\(S/N) cancels out the category of the VP, which is S/N, while *panti-rni* is dependent on *ka*. Hence, we have that ~~S/N~~ ~~(S/N)~~\(S/N) ≡ *ka* (*panti-rni* *). Thereafter, *ka* with the category S/N will cancel out the category N of the subject *ngarrka-ngku*, while *ngarrka-ngku* is dependent on *ka*. Therefore, we have that S/~~N~~ ~~N~~ ≡ *ka*(* *ngarrka-ngku*). All these formulations are in sync with *The Correspondence Principle*.

Equipped with this understanding, we shall now turn to our first example (that is, the German sentence (10)) to be illustrated below. The major steps are shown for the remaining examples.

(10) *Sie haben den versucht zu lesen.*

they have it tried to read

'They have tried to read it.'

(10a) *A CG derivation in the phrase structure tree (PSG → CG)*

Figure 16 depicts the CG derivation of (10) in its PSG tree and Figure 17 depicts the CG derivation of (10).

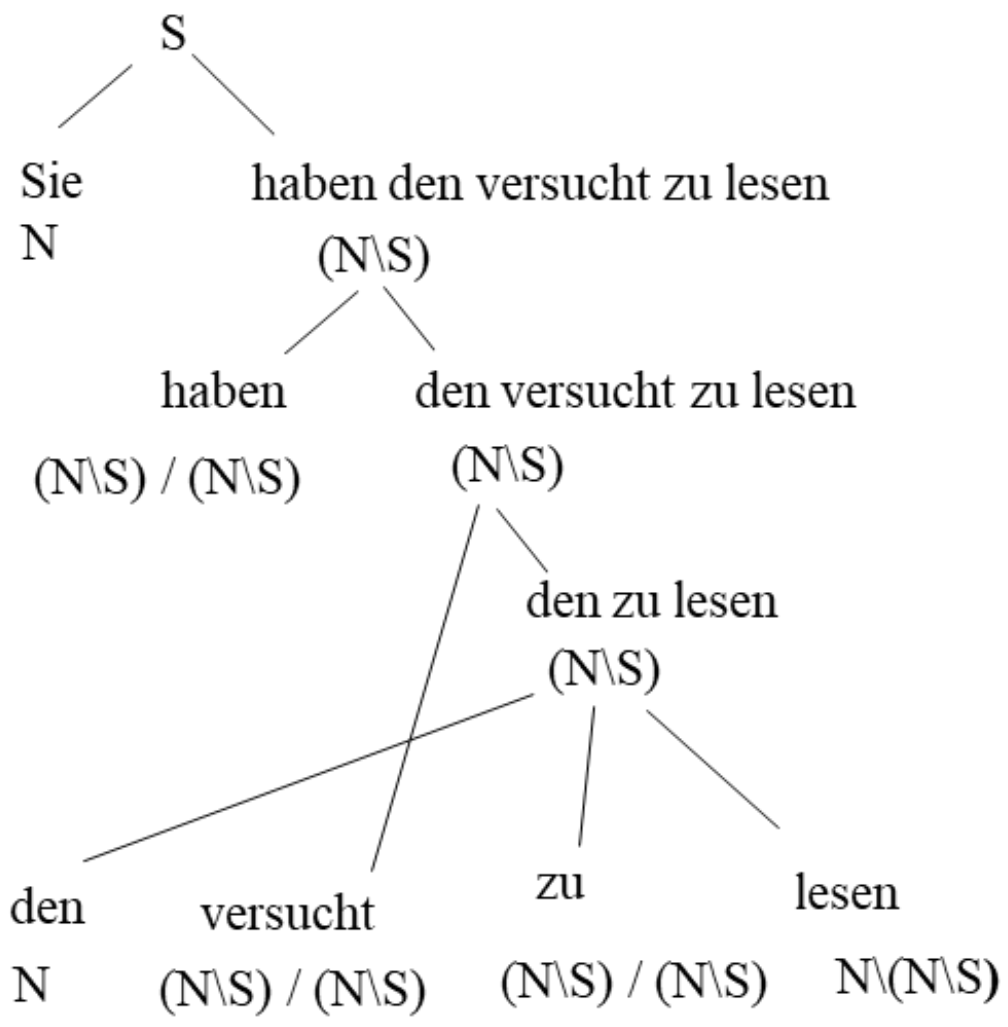

Fig 16. The CG derivation in a PSG tree

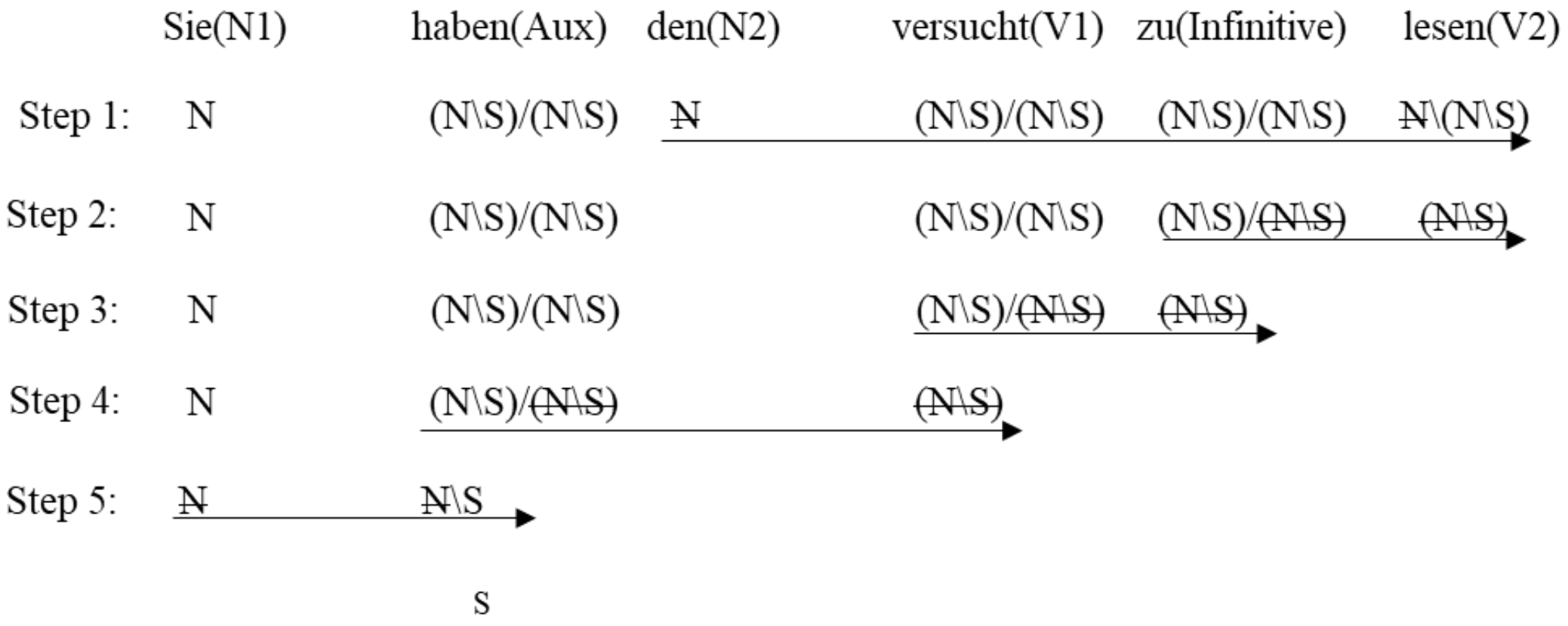

Fig 17. The CG derivation of (10)

*The illustration of Fig 17:*

➔ In step1, the category of *den* is cancelled out with respect to the category of *lesen*.

➔ In step 2, the output of step 1 (category of *den lesen*) becomes the input to step 2 and is cancelled out with respect to the category of *zu*.

➔ In step 3, the output of step 2 (category of *den zu lesen*) now becomes the input to step 4 and is cancelled out with respect to the category of *versucht*.

➔ In step 4, the output of step 3 (category of *den versucht zu lesen*) becomes the input to step 5 and is cancelled out with respect to the category of *haben*.

➔ In step 5, the output of step 4 (category of *haben den versucht zu lesen*) is cancelled with respect to the category of *Sie*, resulting in the final category S.

It may be noted that though it has been mentioned previously that standard PSG trees do not allow for criss-crossing lines in the tree diagrams, the crossing lines are drawn in order to explain how the cancellation of categorial functions can be implemented with the help of the PSG tree as seen in Figure 16. The cancellation of arguments of a function proceeds in accordance with the constituency relations in PSG[6], as seen in Figure 17. The exact manner in which tree branches are or can be tangled reflects the way categorial derivations can work, thus uniting CG derivations with PSG. That the crossing lines are made insignificant in CG is substantiated by the specification of the series of steps for the categorial derivation of the sentence which is shown right below the corresponding tree diagram (Figure 16).

(10b) *Dependency functions in terms of CG formulae (DG → CG)*

The dependency graph for the sentence (10) is shown below.

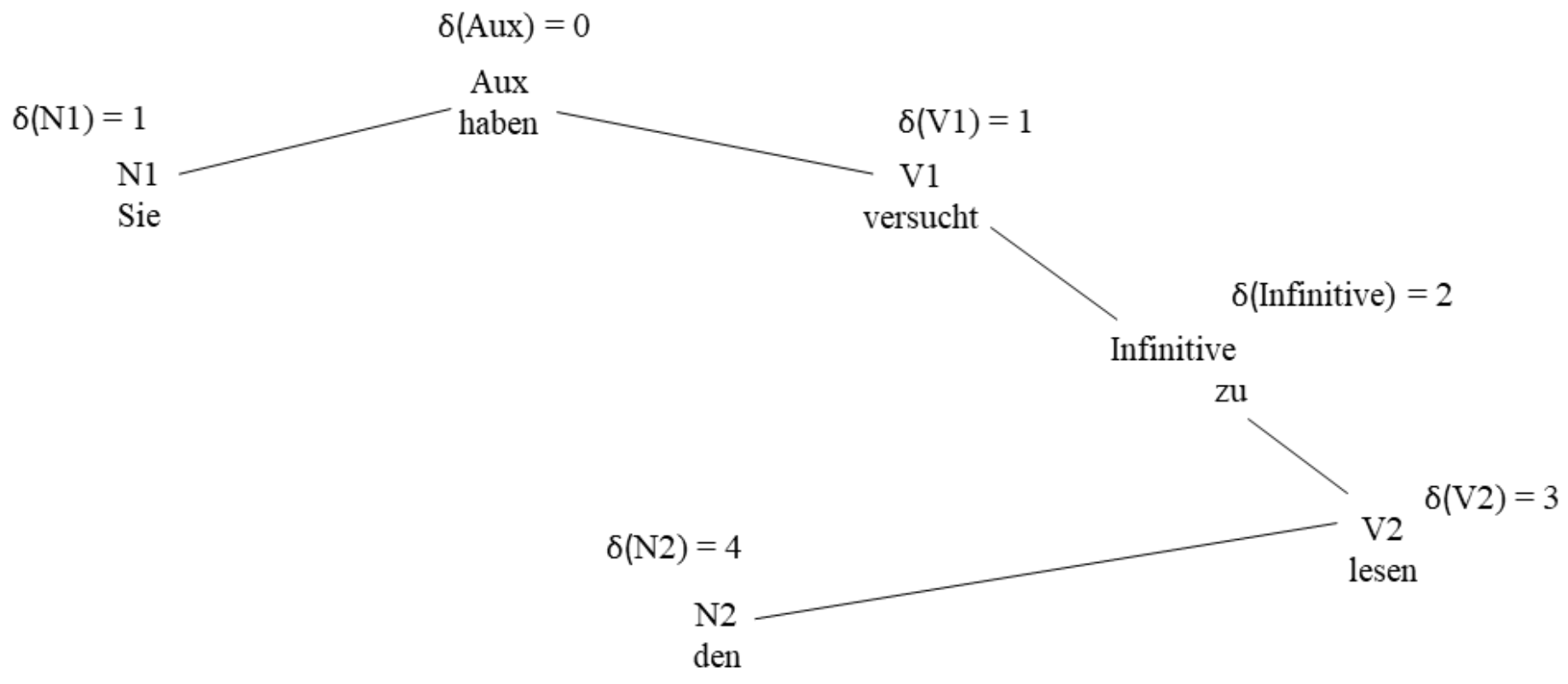

Fig 18. A DG graph corresponding to (10)

[6] In this regard, one may also take note of the treatment of verbal complexes in German with discontinuous constituents for a different analysis (see Müller 2002: 136–138, 154–15).

The above figure illustrates the following dependencies:

i. *den* is dependent on *lesen*.

ii. *lesen* is dependent on *zu*.

iii. *zu* is dependent on *versucht*.

iv. *versucht* is dependent on *haben*.

v. *Sie* is dependent on *haben*.

Based on Figures 17 and 18, the dependency functions capturing all the functor-argument relations/cancellations can be formulated as follows:

Step 1: ~~δ(N2)~~ ~~δ(N2)~~\ δ(V2)

This step captures the functor-argument relation between *lesen* and *den* corresponding to the step 1 of the CG derivation in Figure 17. This step builds the meaning conveyed through *read it*.

Step 2: δ(Infinitive)/~~δ(V2)~~ ~~δ(V2)~~

This step captures the functor-argument relation between *zu* and *lesen* corresponding to step 2 of the CG derivation in Figure 17. This step builds the meaning conveyed through *to read it*.

Step 3: δ(V1)/~~δ(Infinitive)~~ ~~δ(Infinitive)~~

This step captures the functor-argument relation between *zu* and *versucht* corresponding to step 3 of the CG derivation in Figure 17. This step builds the meaning conveyed through *tried to read it*.

Step 4: δ(Aux)/~~δ(V1)~~ ~~δ(V1)~~

This step captures the functor-argument relation between *haben* and *versucht* corresponding to step 4 of the CG derivation in Figure 17. This step builds the meaning conveyed through *have tried to read it*.

Step 5: ~~δ(N1)~~ ~~δ(N1)~~\δ(Aux)

This step captures the functor-argument relation between *Sie* and *haben* corresponding to step 5 of the CG derivation in Figure 17. This step builds the meaning conveyed through *They have tried to read it*.

(10c) *CG → DG derivation*

Here we use *the correspondence principle* and show how the dependency relations can be derived from the categories assigned to the words and the subsequent CG derivation. When each step of the CG derivation is taken into account and expressed in terms of head-dependent relations, the corresponding dependency relation between the functor and the argument can be established.

In this derivation, the equivalence relation is established from RHS to LHS. Hence the CG derivation would be the starting point. The aim is to establish a DG relation for each step of the CG derivation. In other words, for every functor-argument relation, the equivalent head-dependent relation is to be established. Finally, by considering all the head-dependent relations that are established from the CG derivation and other possible head-dependent relations (if any), the DG graph of the sentence can be drawn.

*Step 1: The CG relation between 'lesen' and 'den'*

In this CG relation, *lesen* (V2) is the functor and *den*(N2) is the argument. The direction of the argument is to the left. If we consider *lesen* (V2) to be A and *den*(N2) to be B, the RHS would be N2\V2 or *den\lesen* or B\A. There is a direct dependency relation between the functor and the argument - *lesen* and *den*, with *lesen* as the head and *den* as its dependent. Accordingly, the LHS would be V2(N2*) or *lesen*(*den**) or A(B*). Thus the equivalence relation for this CG relation would be:

V2(N2*) ≡ N2\V2

*Step 2: The CG relation between 'lesen' and 'zu'*

In this CG relation, *zu*(Infinitive) is the functor and *lesen*(V2) is the argument. The direction of the argument is to the right. If we consider *zu* (Infinitive) to be A and *lesen* (V2) to be B, the RHS would be Infinitive/V2 or *zu/lesen* or A/B. There is a direct dependency relation between the functor and the argument - *zu* and *lesen*, with *zu* as the head and *lesen* as its dependent. Accordingly, the LHS would be *zu*(**lesen*) or Infinitive(*V2). Thus the equivalence relation for this CG relation would be:

Infinitive(*V2) ≡ Infinitive/V2

*Step 3: The CG relation between 'versucht' and 'zu'*

In this CG relation, *versucht* (V1) is the functor and *zu* (Infinitive) is the argument. The direction of the argument is to the right. If we consider *versucht*(V1) to be A and *zu* (Infinitive) to be B, the RHS would be V1/Infinitive or *versucht/zu* or A/B. There is a direct dependency relation between the functor and the argument - *versucht* and *zu*, with *versucht* as the head and *zu* as its dependent. Accordingly, the LHS would be *versucht*(**zu*) or V1(*Infinitive) or A(*B). Thus the equivalence relation for this CG relation would be:

V1(*Infinitive) ≡ V1/Infinitive

*Step 4: The CG relation between 'haben' and 'versucht'*

In this CG relation, *haben*(Aux) is the functor and *versucht*(V1) is the argument. The argument is to the right of the functor. If we consider *haben*(Aux) to be A and *versucht*(V1) to be B, the RHS would be *haben*/*versucht* or Aux/V1 or A/B. There is a direct dependency relation between the functor and the argument – *haben* and *versucht*, with *haben* as the head and *versucht* as its dependent. Accordingly, the LHS would be *haben*(**versucht*) or Aux(*V1) or A(*B). Thus the equivalence relation for this CG relation would be:

Aux(*V1) ≡ Aux/V1

*Step 5: The CG relation between 'Sie' and 'haben'*

In this CG relation, *haben* (Aux) is the functor and *Sie*(N1) is the argument. The direction of the argument is to the left. If we consider *Sie*(N1) to be B, the RHS would be N1\Aux or *Sie**haben* or B\A (A = *haben*). There is a direct dependency relation between the functor and the argument – *haben* and *Sie*, with *haben* as the head and *Sie* as its dependent. Accordingly, the LHS would be *haben*(*Sie**) or Aux(N1*) or A(B*). Thus the equivalence relation for this CG relation would be:

Aux(N1*) ≡ N1\Aux

Thus, combining the DG relations of all the steps, we get the following.

i. V2(N2*) or *den* is dependent on *lesen*
ii. Infinitive(*V2) or *lesen* is dependent on *zu*
iii. V1(*Infinitive) or *zu* is dependent on *versucht*
iv. Aux(*V1) or *versucht* is dependent on *haben*
v. Aux(N1*) or *Sie* is dependent on *haben*

Based on the above dependency relations, we arrive at the DG graph, that is, Figure 18.

(10d) *CG → PSG derivation*

In this German sentence, discontinuity arises because *lesen* and *den* are not contiguous in the linear order of the sentence. However, in a PSG tree the cancellation of arguments proceeds as per the constituent structure.

Step 1 of the CG derivation would mean that N (*den*) and V (*lesen*) form a single constituent VP *den lesen* with V as its head. This corresponds to the output *read it* as seen in Figure 19.

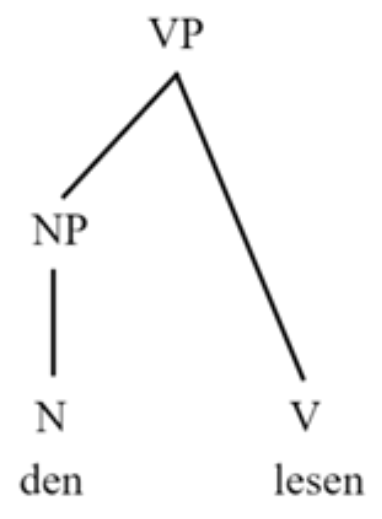

Fig 19. The PSG tree corresponding to step 1 of the CG derivation

Step 2 of the CG derivation would correspond to the output *to read it* which forms an IP constituent *den zu lesen* with I as its head as seen in Figure 20.

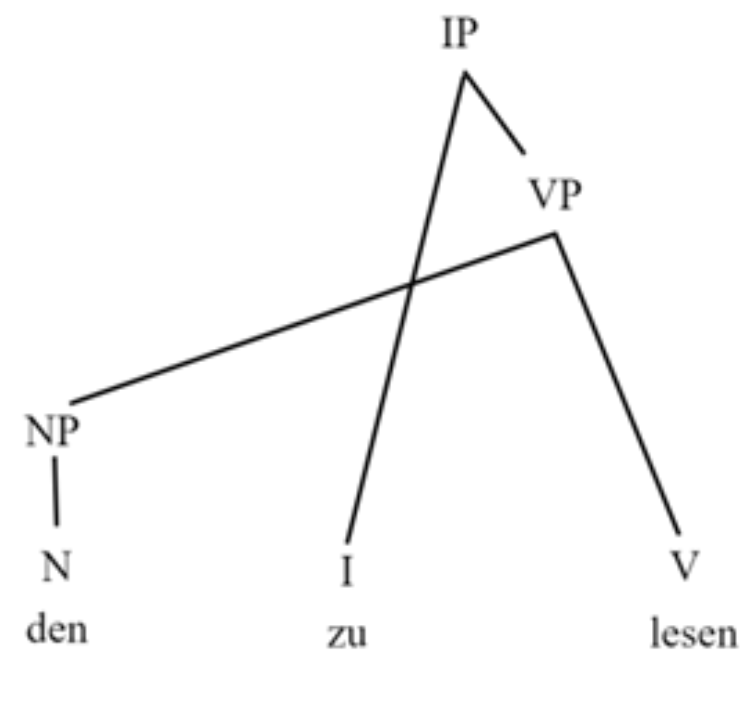

Fig 20. The PSG tree corresponding to step 2 of the CG derivation

Step 3 of the CG derivation would correspond to the output *tried to read it* which forms the VP constituent *den versucht zu lesen*, as seen in Figure 21.

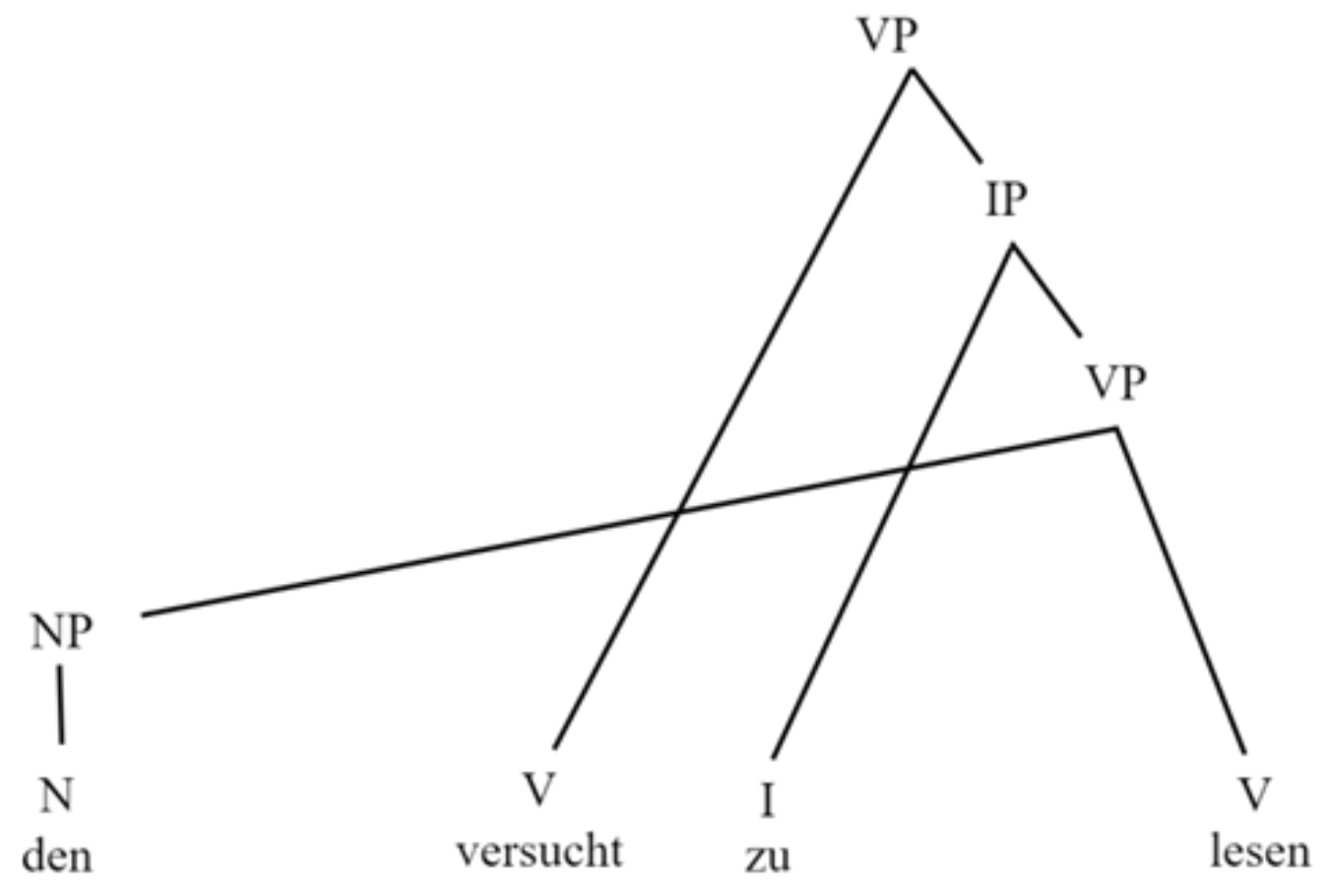

Fig 21. The PSG tree corresponding to step 3 of the CG derivation

Step 4 of the CG derivation would correspond to the output *have tried to read it* which forms the AuxP constituent *haben den versucht zu lesen* as seen in Figure 22.

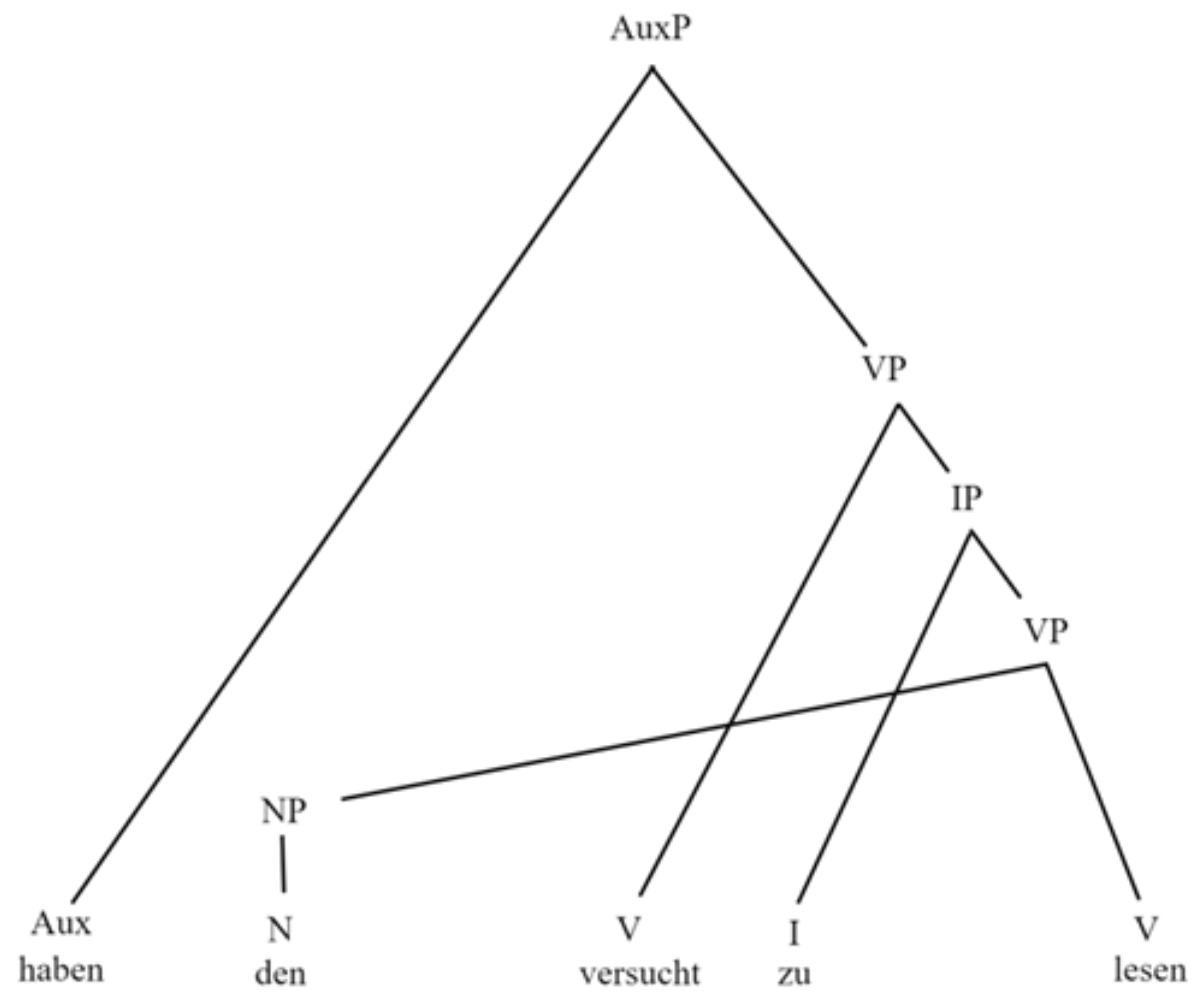

Fig 22. The PSG tree corresponding to step 4 of the CG derivation

After all the arguments are cancelled out in Step 5, 'S' is the final output. This corresponds to the tangled tree diagram Figure 23.

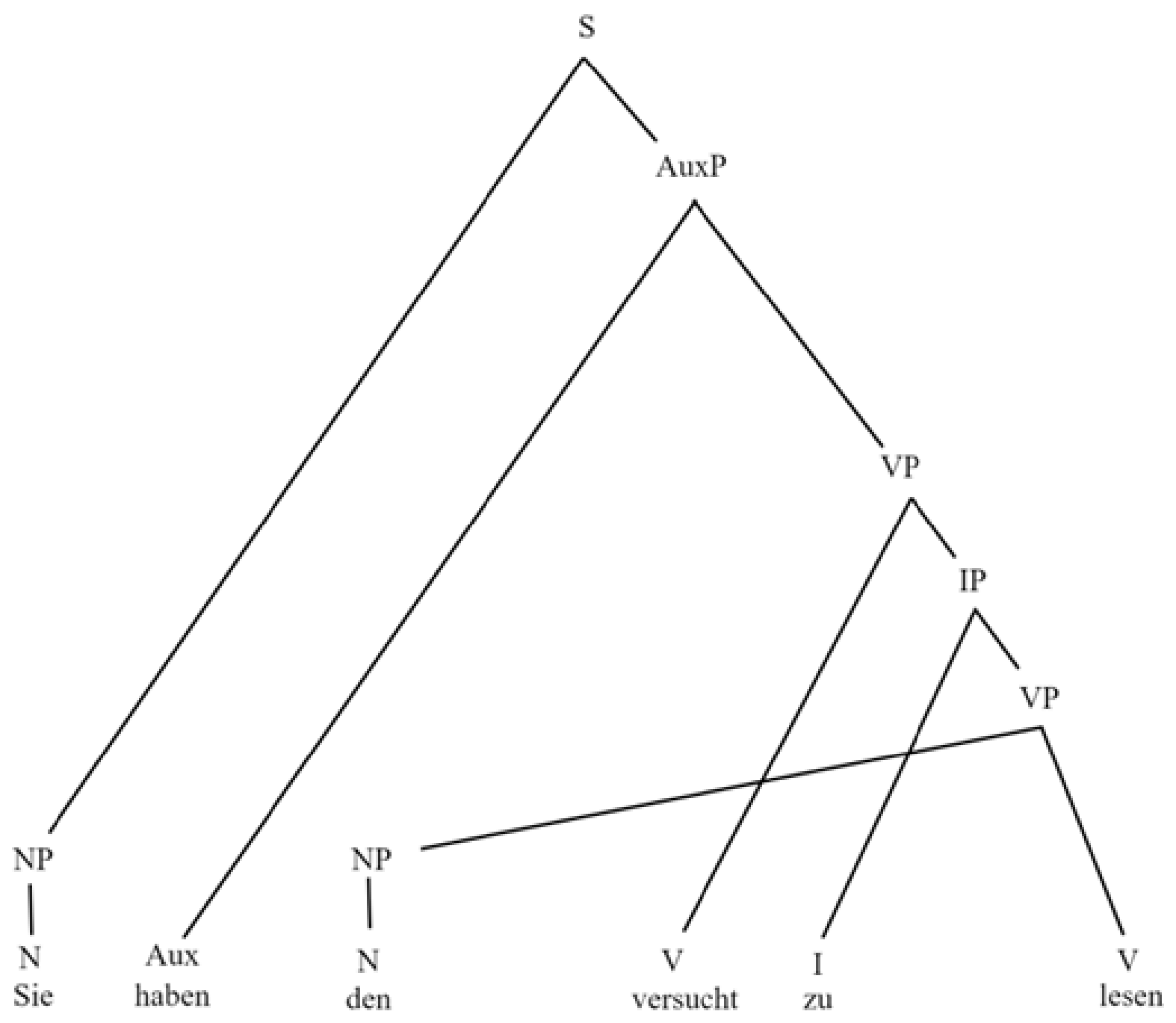

Fig 23. The final PSG tree

Accordingly, the corresponding PSG rules are as follows.

S → NP AuxP

AuxP → Aux VP

VP → V IP

IP → I VP

VP → NP V

NP → N

(10e) *A unified representation (from DG to CG to PSG)*

This is depicted in Figure 24.

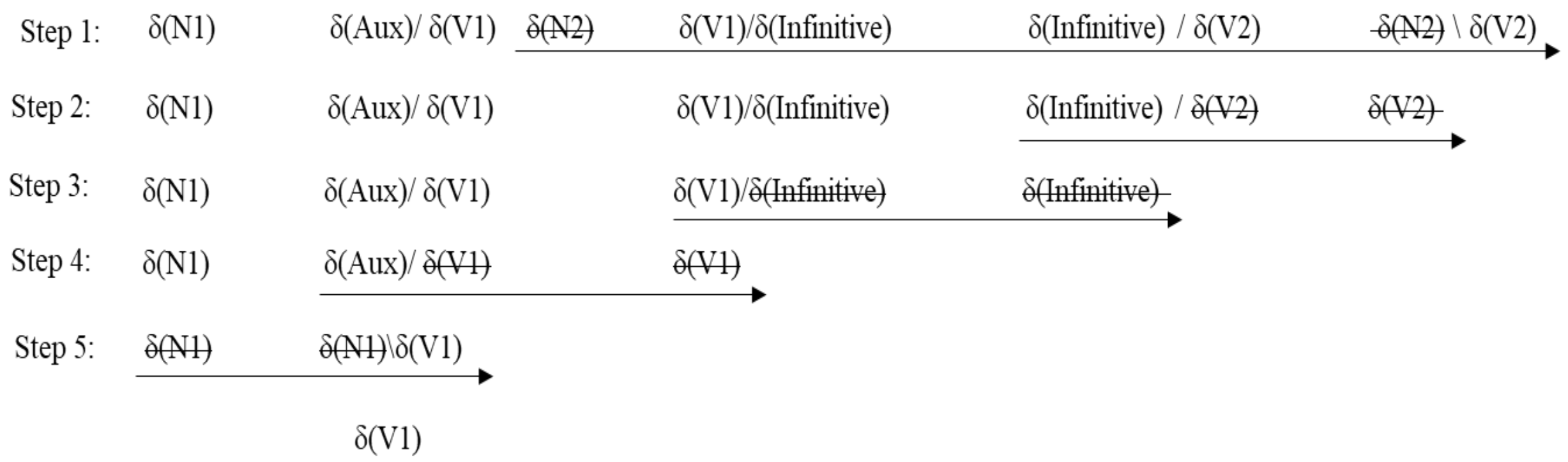

Fig 24. A unified representation

We may now turn to our second example. Consider the following Croatian sentence (11) (see Van Valin 2001, 88):

(11) *Naša* *je* *učionica* *udobna*.

our is classroom comfortable

'Our classroom is comfortable.' (Van Valin 2001:88)

(11a) *A CG derivation in the phrase structure tree (PSG → CG)*

Figure 25 depicts the CG derivation of (11) in its PSG tree and Figure 26 depicts the CG derivation of (11).

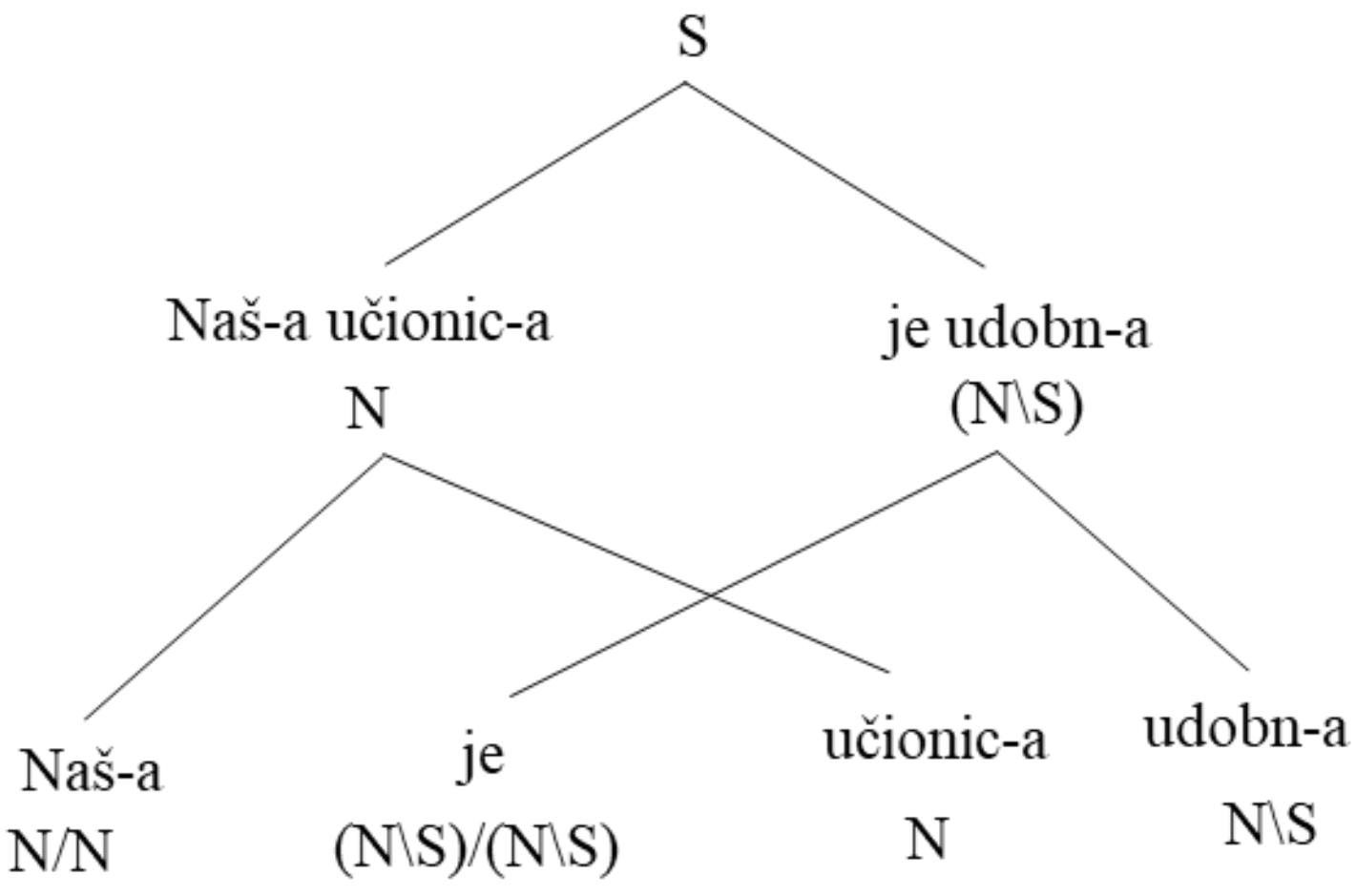

Fig 25. The CG derivation in PSG tree

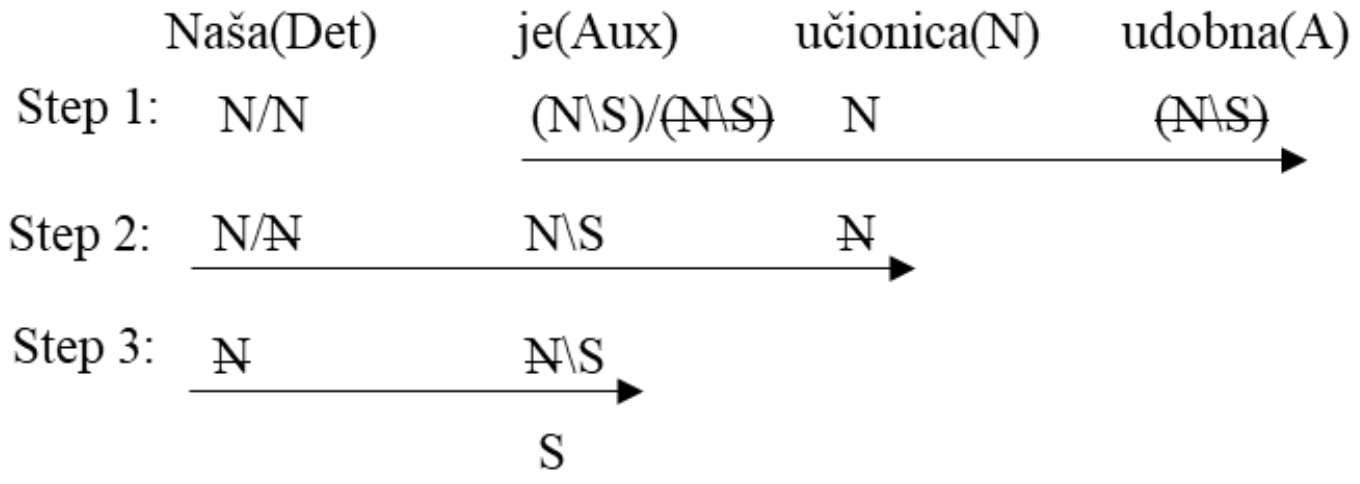

Fig 26. The CG derivation of (11)

(11b) *Dependency functions in terms of CG formulae (DG → CG)*

The dependency graph for the sentence (11) is shown below.

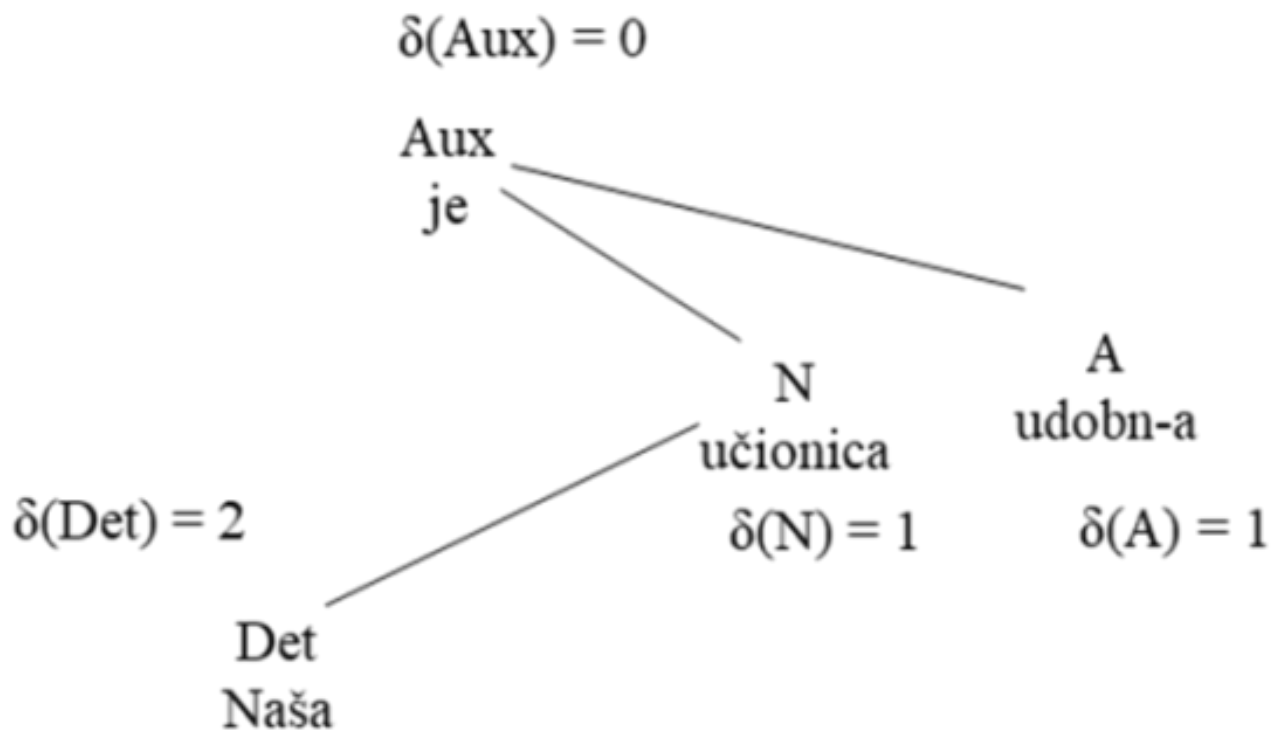

Fig 27. A dependency graph of (11)

The following steps correspond to those in the CG derivation in Figure 26.

Step 1: δ(Aux)/δ~~(A)~~ ~~δ(A)~~

Step 2: δ(Det)/~~δ(N)~~ ~~δ(N)~~

Step 3: δ(Aux)/ ~~δ(N)~~ ~~δ(N)~~

Note: In step 3, by using the correspondence principle, we have that *je*(**učionica*) ≡ *Naša**je*.

In other words, Aux(*N) ≡ Det\Aux. We can also express it as: A(*B) ≡ B\A. Since *Naša* and *je* do not participate in any (direct) dependency relation as seen in Figure 27, the functor-argument relation is constructed through Aux and N in step 3.

(11c) *CG → DG derivation*

The steps formulated below correspond to those in Fig 26.

*Step 1: The CG relation between 'je' and 'udobna'*

In this CG relation, *je* (Aux) is the functor and *udobna*(Adj) is the argument. The direction of the argument is to the right. If we consider *je* (Aux) to be A and *udobna*(Adj) to be B, the RHS would be Aux/Adj or *je*/*udobna* or A/B. There is a direct dependency relation between the functor and the argument - *je* and *udobna*, with *je* as the head and *udobna* as its dependent. Accordingly, the LHS would be *je*(**udobna*) or Aux(*Adj) or A(*B). Thus the equivalence relation for this CG relation would be: Aux(*Adj) ≡ Aux/Adj.

*Step 2: The CG relation between 'Naša' and 'učionica'*

In this CG relation, *Naša*(Det) is the functor and *učionica* (N) is the argument. The direction of the argument is to the right. If we consider *učionica* (N) to be A and *Naša*(Det) to be B, the RHS would be Det/N or *Naša*/*učionica* or B/A. There is a direct dependency relation between the functor and the argument - *Naša* and *učionica*, with *učionica* as the head and *Naša* as its dependent. Accordingly, the LHS would be the following: *učionica* (*Naša**) or N(Det*) or A(B*). Thus the equivalence relation for this CG relation would be: N(Det*) ≡ Det/N.

*Step 3: The CG relation between 'Naša' and 'je'*

In this CG relation, *je* (Aux) is the functor and *Naša* (Det) is the argument. The direction of the argument is to the left. If we consider *je*(Aux) to be A, the RHS would be Det\Aux or *Naša\je* or B\A (B = Det). However, there is no direct dependency relation between the functor and the argument - *je* and *Naša*. Rather, *učionica* (N) is dependent on *je*(Aux). In other words, Aux(*N) or *je*(**učionica*) or A(*B) [B = *učionica* ] would be the LHS. Since the functor and the argument do not participate in a direct head and dependent relationship, considering *je* to be A on the RHS would implicitly indicate that B on the RHS is its argument *Naša* and B on the LHS is its dependent *učionica* .Thus the equivalence relation for this CG relation would be: Aux(*N) ≡ Det\Aux. This clearly shows that *je* (Aux) is the head of *učionica* (N) but is the functor of the argument *Naša* (Det). Thus, combining the DG relations of all the steps would give us the following.

(i) Aux(*Adj) or *udobna* is dependent on *je*

(ii) N(Det*) or *Naša* is dependent on *učionica*

(iii) Aux(*N) or *učionica* is dependent on *je*

Based on the above dependency relations, we arrive at the DG graph, that is, Figure 27.

(11d) *CG → PSG derivation*

In this Croatian sentence, discontinuity arises because (i) *Naša* and *učionica* and (ii) *je* and *udobna* are not contiguous in the linear order of the sentence. However, in a PSG tree, the cancellation of arguments proceeds as per the constituent structure.

Step1 of the CG derivation would mean that Aux and AP form a single constituent AuxP with Aux as the head. The analysis of *udobna* as a separate constituent 'AP' is drawn from the fundamental principles of PSG as seen in Figure 28.

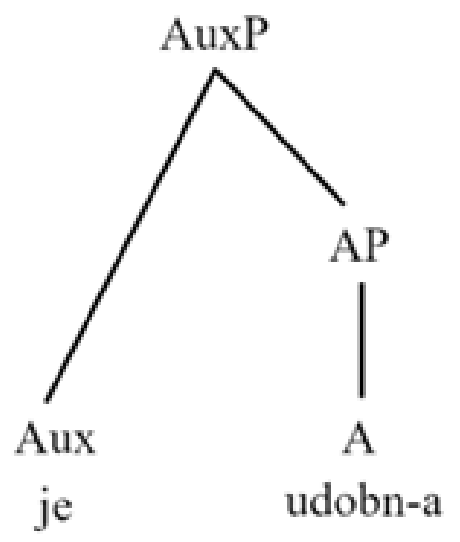

Fig 28. The PSG tree corresponding to step 1 of the CG derivation

Step 2 of the CG derivation would mean that Det and N form a single constituent NP with N as the head as seen in Figure 29.

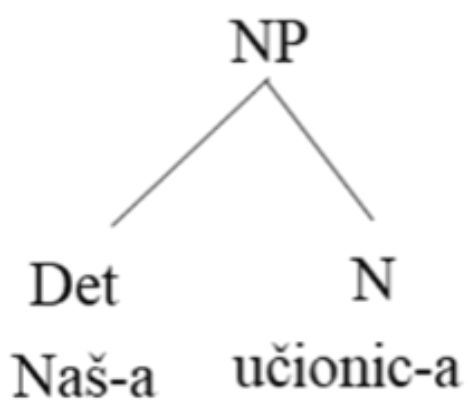

Fig 29. The PSG tree corresponding to step 2 of the CG derivation

Step 3 of the CG derivation indicates that NP and AuxP form a single constituent S. Hence, in this step the remaining arguments are cancelled out resulting in the final output S. The corresponding tangled diagram of the sentence is shown below in Figure 30. Finally, Figure 31 depicts a unified representation.

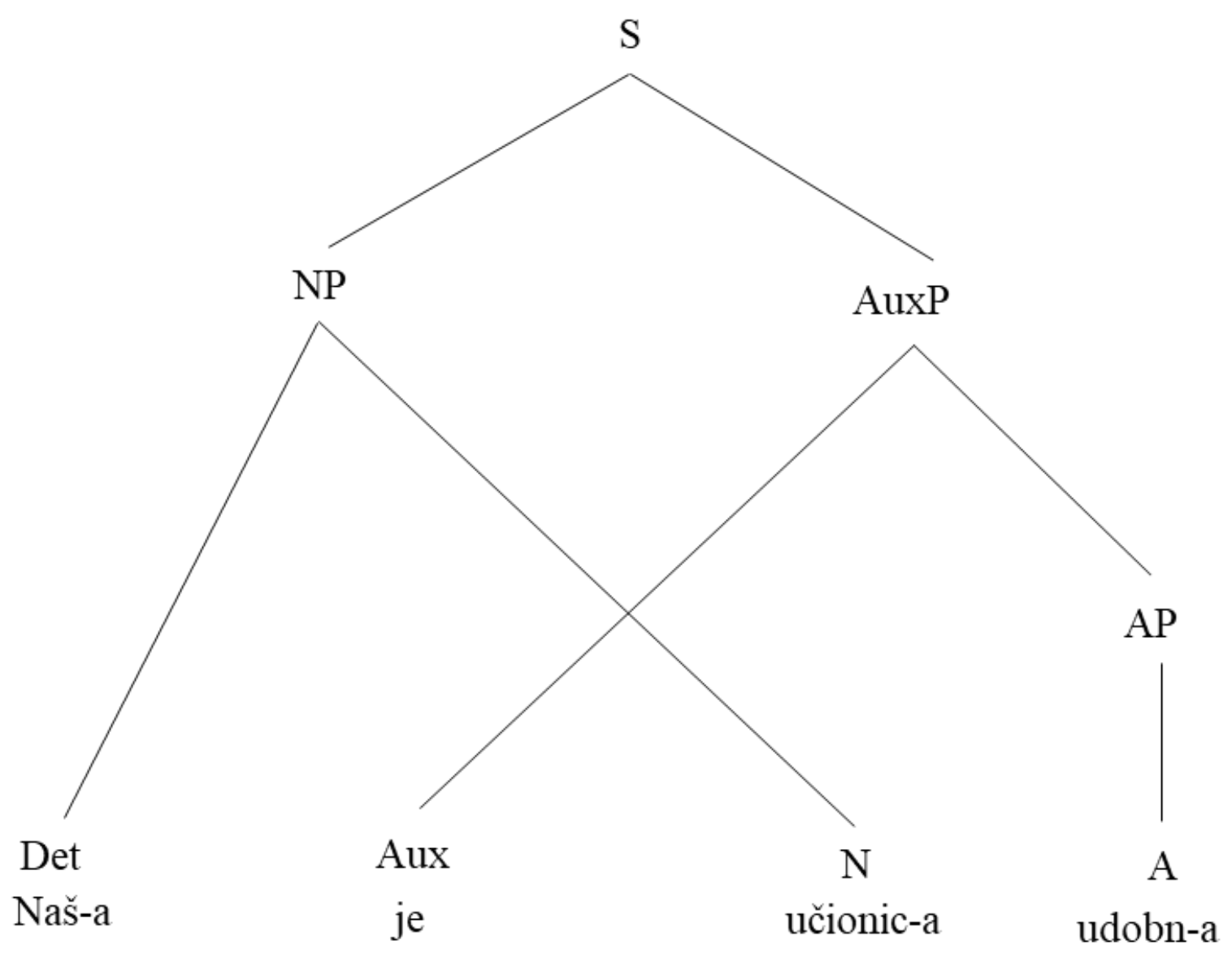

Fig 30. The final PSG tree

<u>The corresponding PSG rules are:</u>

S → NP AuxP

NP → Det N

AuxP → Aux AP

AP → A

(11e) *A unified representation (from DG to CG to PSG)*

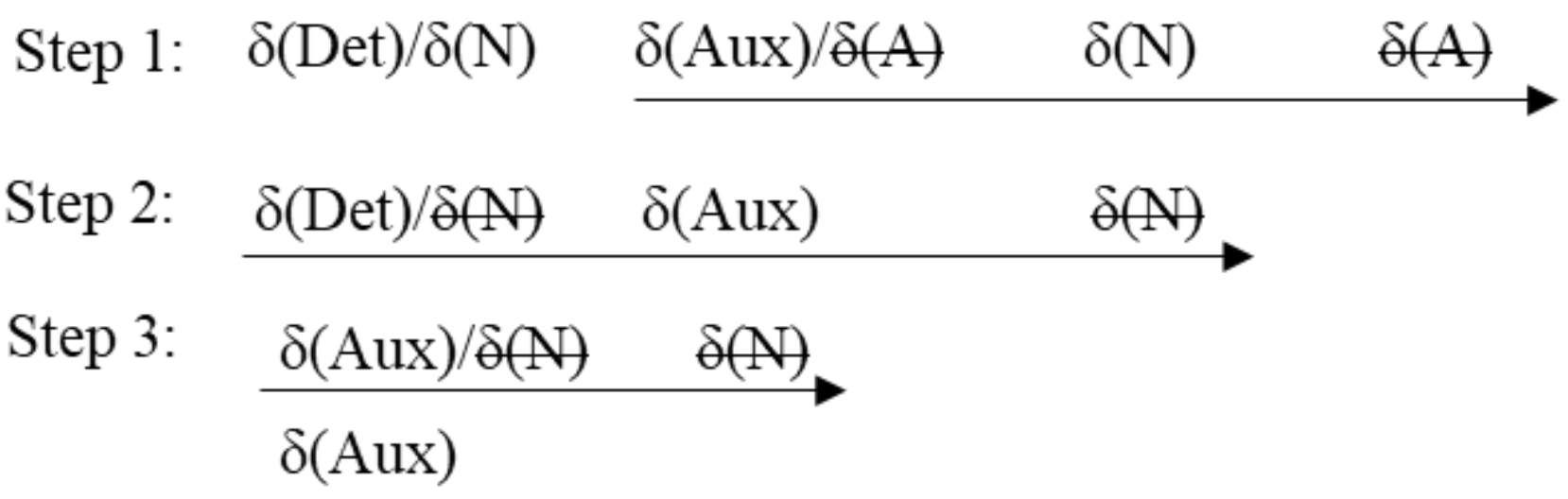

Fig 31. A unified representation

Now, it is time to consider our third example (12) from Kalkatungu.

(12) *N̠a-ci kuḷa-ji l̠aji t̠uar-Ø mal̠t̠a-Ø japacara- t̠u.*

1SG-DAT father-ERG kill snake-ABS mob-ABS clever-ERG

‘My clever father killed the snakes.[7]’ (Van Valin 2001:88)

[DAT=dative case marking]

(12a) *A CG derivation in the phrase structure tree (PSG → CG)*

Figure 32 depicts the CG derivation of (12) in its PSG tree and Figure 33 depicts the CG derivation of (12).

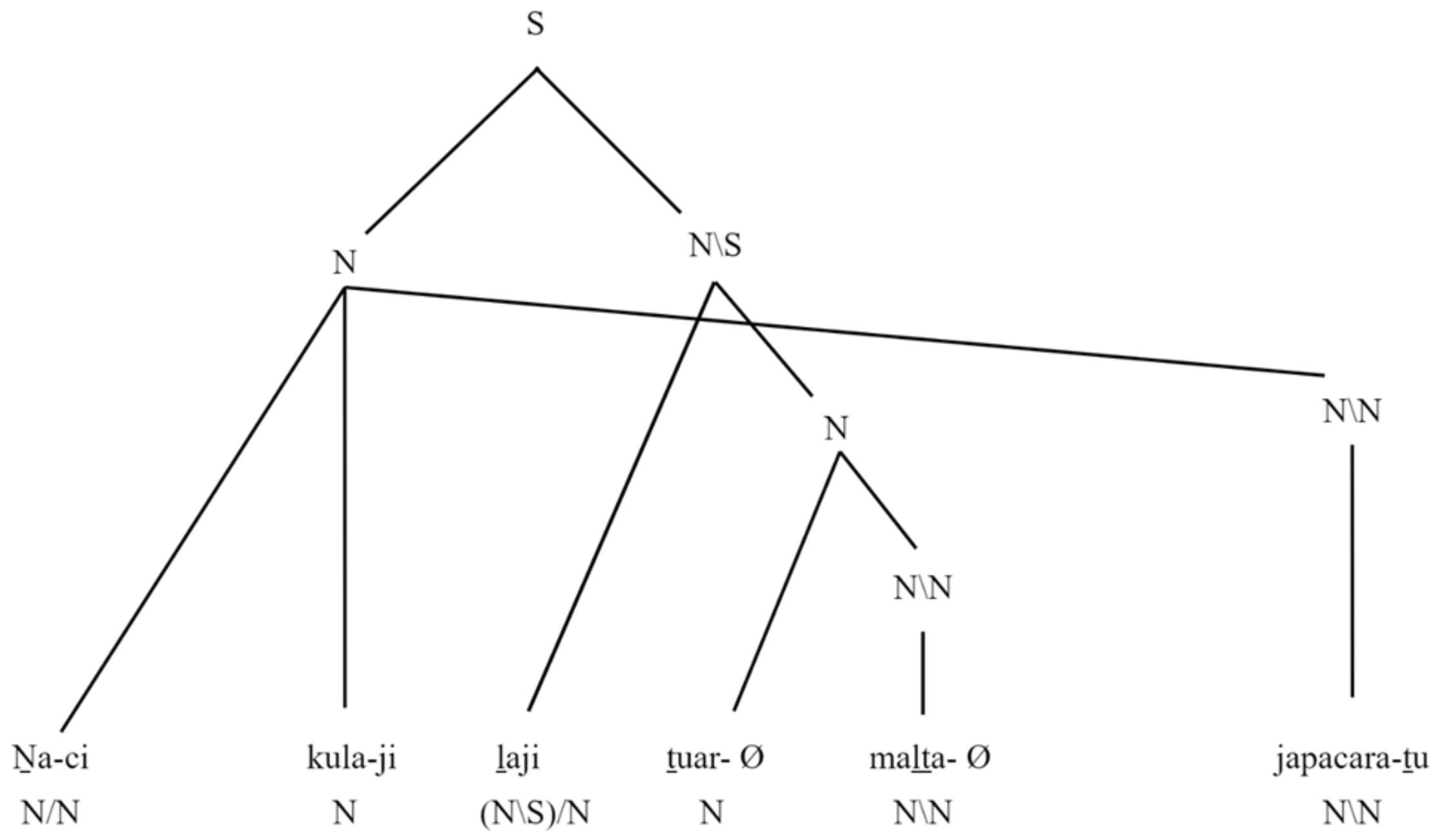

Fig 32. The CG derivation in PSG tree

[7] The literal translation of *t̠uar-Ø mal̠t̠a-Ø* is indeed of ‘the mob of snakes’, but the author in the original source has found it appropriate to simplify it for the non-literal translation.

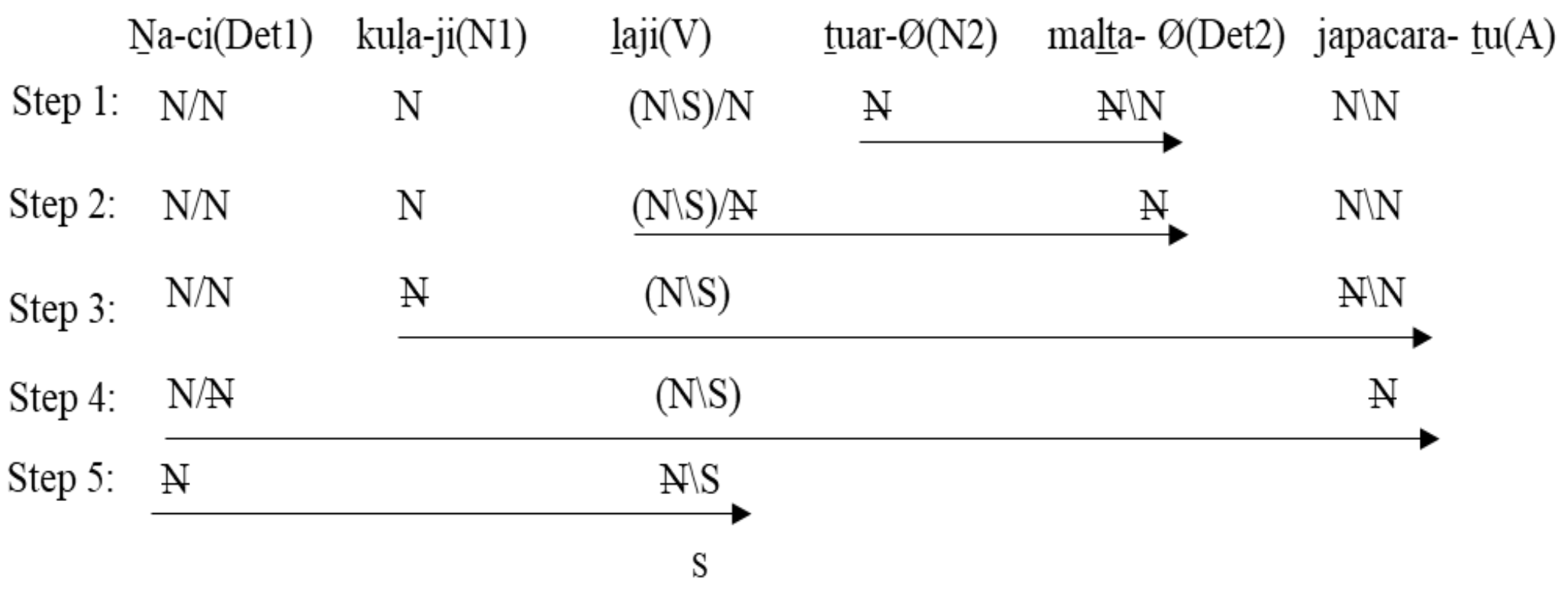

Fig 33. The CG derivation of (12)

(12b) *Dependency functions in terms of CG formulae (DG → CG)*

The dependency graph for the sentence (12) is shown below.

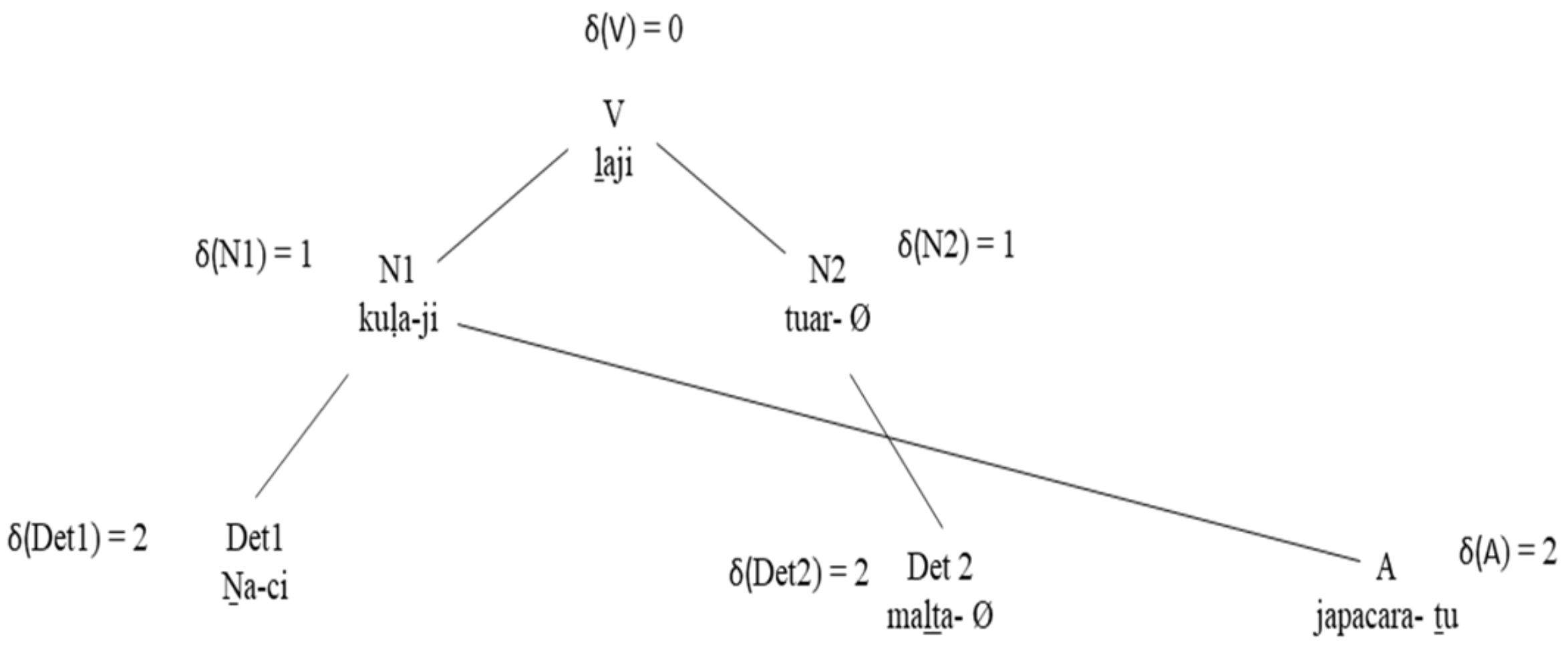

Fig 34. A dependency graph of (12)

The following steps correspond to those in the CG derivation in Fig 33 in the same order.

Step 1: ~~δ(N2)~~ ~~δ(N2)~~\-δ(Det2)

Step 2: δ(V)/ ~~δ(N2)~~ ~~δ(N2)~~

Step 3: ~~δ(N1)~~ ~~δ(N1)~~\ δ(A)

Step 4: δ(Det1)/~~δ(N1)~~ ~~δ(N1)~~

Step 5: ~~δ(N1)~~ ~~δ(N1)~~\δ(V)

Note (based on Figure 34):

(i) In step 2, by using the correspondence principle, we have that *l̲aji*(**t̲uar-Ø*) ≡ *l̲aji*/*mal̲t̲a-Ø*. In other words, V(*N2) ≡ V/Det2. We can also express it as: A(*B) ≡ A/B. Since *mal̲t̲a-Ø* and *l̲aji* do not participate in any (direct) dependency relation, the functor-argument relation is constructed through V and N2 in step 2.

(ii) In step 5, by using the correspondence principle, we have that *l̲aji*(*kuḷa-ji**) ≡ *N̲a-ci**l̲aji*. In other words, V(N1*) ≡ Det1\V. We can also express it as: A(B*) ≡ B\A. Since *N̲a-ci* and *l̲aji* do not participate in any (direct) dependency relation, the functor-argument relation is constructed through V and N1 in step 5.

(12c) *CG → DG derivation*

The steps formulated below correspond to those in Fig 33.

*Step 1: The CG relation between 't̲uar-Ø' and 'mal̲t̲a-Ø '*

In this CG relation, *mal̲t̲a-Ø* (Det2) is the functor and *t̲uar-Ø* (N2) is the argument. The direction of the argument is to the left. If we consider *t̲uar-Ø* (N2) to be A and *mal̲t̲a-Ø* (Det2) to be B, the RHS would be N2\Det2 or *t̲uar-Ø*\ *mal̲t̲a-Ø* or A\B. There is a direct dependency relation between the functor and the argument - *t̲uar-Ø* and *mal̲t̲a-Ø*, with *t̲uar-Ø* as the head and *mal̲t̲a-Ø* as its dependent. Accordingly, the LHS would be *t̲uar-Ø*(**mal̲t̲a-Ø*) or N2(*Det2). Thus the equivalence relation for this CG relation would be: N2(*Det2) ≡ N2\Det2

*Step 2: The CG relation between 'l̲aji' and 'mal̲t̲a-Ø '*

In this CG relation, *ḻaji* (V) is the functor and *maḻṯa-Ø* (Det2) is the argument.The direction of the argument is to the right. If we consider *ḻaji* (V) to be A, the RHS would be *ḻaji*/ *maḻṯa-Ø* or V/Det2 or A/B (B = *maḻṯa-Ø*). There is no direct dependency relation between the functor and the argument – *ḻaji* (V) and *maḻṯa-Ø*, rather *ṯuar-Ø* (N2) is dependent on *ḻaji* (V). In other words, *ḻaji*(**ṯuar-Ø*) or V(*N2) or A(*B) [B = *ṯuar-Ø*] would be the LHS. Since they do not participate in any direct head and dependent relationship, considering *ḻaji* (V) to be A on the RHS would implicitly indicate that B on the RHS is its argument *maḻṯa-Ø* and B on the LHS is its dependent *ṯuar-Ø*. Thus the equivalence relation for this CG relation would be: V(*N2) ≡ V/Det2. This clearly shows that *ḻaji* (V) is the head of *ṯuar-Ø* (N2) and the functor of the argument *maḻṯa-Ø* (Det2).

*Step 3: The CG relation between 'kuḷa-ji' and 'japacara-tu'*

In this CG relation, *japacara-tu* (Adj) is the functor and *kuḷa-ji* (N1) is the argument. The direction of the argument is to the left. If we consider *japacara-tu*(Adj) to be B and *kuḷa-ji* (N1) to be A, the RHS would be N1\Adj or *kuḷa-ji\japacara-tu* or A\B. There is a direct dependency relation between the functor and the argument – *japacara-tu* and *kuḷa-ji* with *kuḷa-ji* as the head and *japacara-tu* as its dependent. Accordingly, the LHS would be *kuḷa-ji* (**japacara-tu*) or N1(*Adj). Thus the equivalence relation for this CG relation would be: N1(*Adj) ≡ N1\Adj. This clearly shows that *kuḷa-ji*(N1) is the head of *japacara-tu* (Adj) but is the argument of the functor *japacara-tu*[8].

*Step 4: The CG relation between 'Ṉa-ci' and 'japacara-tu'*

---

[8] One may wonder why a DP (Determiner Phrase) analysis for NPs (Noun Phrases) is absent, especially in cases like this. While we acknowledge that the DP analysis aligns with the head-modifier relationship, with the determiner functioning as a functor that cancels the noun head's category, we have opted for NP analyses throughout the paper to maintain simplicity in the PSG notation.

In this CG relation, *N̠a-ci* (Det1) is the functor and *japacara-tu* (Adj) is the argument. The direction of the argument is to the right. If we consider *N̠a-ci* (Det1) to be B, the RHS would be Det1/Adj or B/A (A = *japacara-tu*). However, there is no direct dependency relation between the functor and the argument: *N̠a-ci* (Det1) and *japacara-tu* (Adj). Rather, *N̠a-ci* (Det1) is dependent on *kuḷa-ji*(N1). In other words, N1(Det1*) or *kuḷa-ji*(*N̠a-ci**) or A(B*) [A = *kuḷa-ji*] would be the LHS. Since the functor and the argument do not participate in any direct head and dependent relationship, considering *N̠a-ci* (Det1) to be B on the RHS would implicitly mean that A on the RHS is its argument *japacara-tu* and A on the LHS is its head *kuḷa-ji*. The equivalence relation for this CG relation would be: N1(Det1*) ≡ Det1/Adj. This clearly shows that *N̠a-ci* is dependent on *kuḷa-ji* but is the functor of the argument *japacara-tu*.

*Step 5: The CG relation between 'N̠a-ci' and 'l̠aji'*

In this CG relation, *l̠aji*(V) is the functor and *N̠a-ci*(Det1) is the argument. The direction of the argument is to the left. If we consider *l̠aji* to be A, the RHS would be Det1\V or B\A (B = *N̠a-ci*). However, there is no direct dependency relation between the functor and the argument – *N̠a-ci* and *l̠aji*; rather, *kuḷa-ji* (N1) is dependent on *l̠aji*(V). In other words, V(N1*) or A(B*) [ B = *kuḷa-ji*]. Since the functor and the argument do not participate in a direct head and dependent relationship, considering *l̠aji* (V) to be A on the RHS would implicitly indicate that B on the RHS is its argument *N̠a-ci* and B on the LHS is its dependent *kuḷa-ji*. Thus the equivalence relation for this CG relation would be: V(N1*) ≡ Det1\V. This clearly shows that *l̠aji* (V) is the head of *kuḷa-ji* but is the functor of the argument *N̠a-ci*. Thus combining the DG relations of all the steps:

i. N2(*Det2) or *mal̠t̠a-Ø* is dependent on *t̠uar-Ø*

ii.V(*N2) or *ṯuar-Ø* is dependent on *ḻaji*

iii.N1(*Adj) or *japacara- ṯu* is dependent on *kuḷa-ji*

iv.N1(Det1*) or *Ṉa-ci* is dependent on *kuḷa-ji*

v.V(N1*) or *kuḷa-ji* is dependent on *ḻaji*

Based on these dependency relations, we arrive at the DG graph, that is, Figure 34.

(12d) *CG → PSG derivation*

In this Kalkatungu sentence, discontinuity arises because *kuḷa-ji* and *japacara-tu* are not contiguous in the linear order of the sentence. However, in a PSG tree, cancellation of the arguments proceeds as per the constituent structure.

Step1 of the CG derivation would mean that N (*ṯuar-Ø*) and Det (*malṯa-Ø*) form a single constituent NP with N as the head. This corresponds to the output *the snakes* as seen in Figure 35.

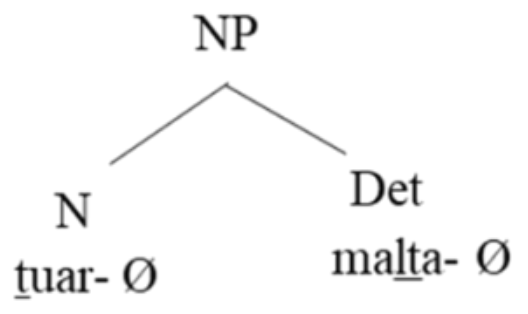

Fig 35. The PSG tree corresponding to step 1 of the CG derivation

Step2 of the CG derivation would correspond to the output *killed the snakes* which forms a VP constituent as seen in Figure 36.

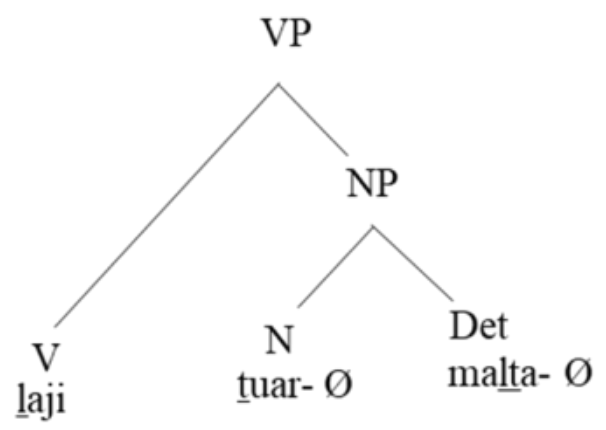

Fig 36. The PSG tree corresponding to step 2 of the CG derivation

Step3 of the CG derivation would correspond to the output *clever father* which forms an NP constituent, as can be seen in Figure 37.

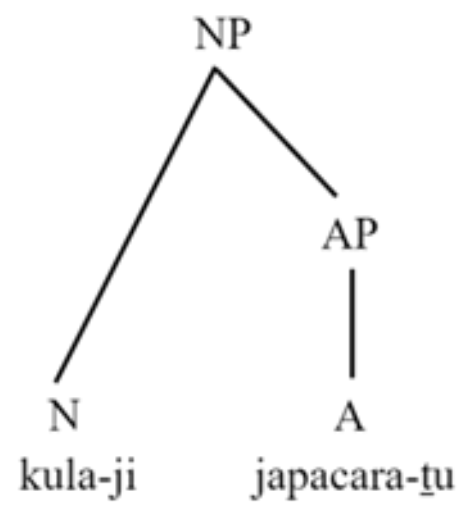

Fig 37. The PSG tree corresponding to step 3 of the CG derivation

Step4 of the CG derivation would correspond to the output *My clever father*. The corresponding PSG tree can be seen in Figure 38.

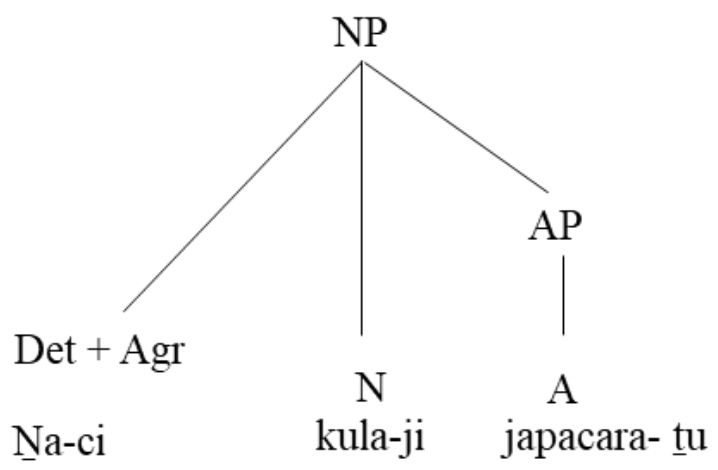

Fig 38. The PSG tree corresponding to step 4 of the CG derivation

In step 5, all the arguments are cancelled out and ‘S’ is the final output. Thus, by using crossed lines, the position of *japacara- ṯu* in the linear order of the sentence can be depicted in the form of a tangled tree as shown below in Figure 39. Finally, Figure 40 depicts the unified representation.

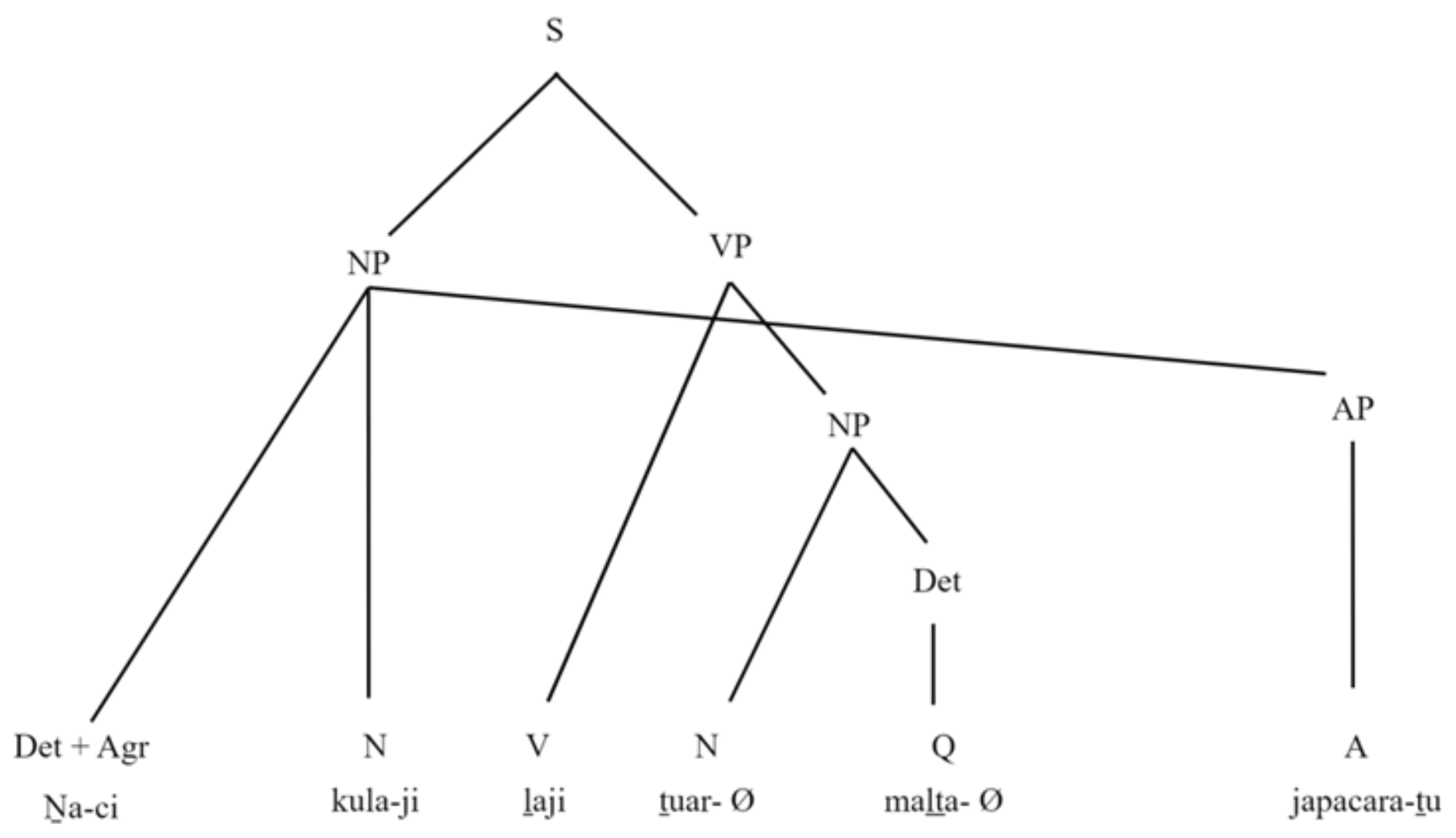

Fig 39. The final PSG tree

Accordingly, the corresponding PSG rules are:

S → NP VP

NP → Det N AP

AP → A

VP → V NP

NP → N Det

(12e) *A unified representation (from DG to CG to PSG)*

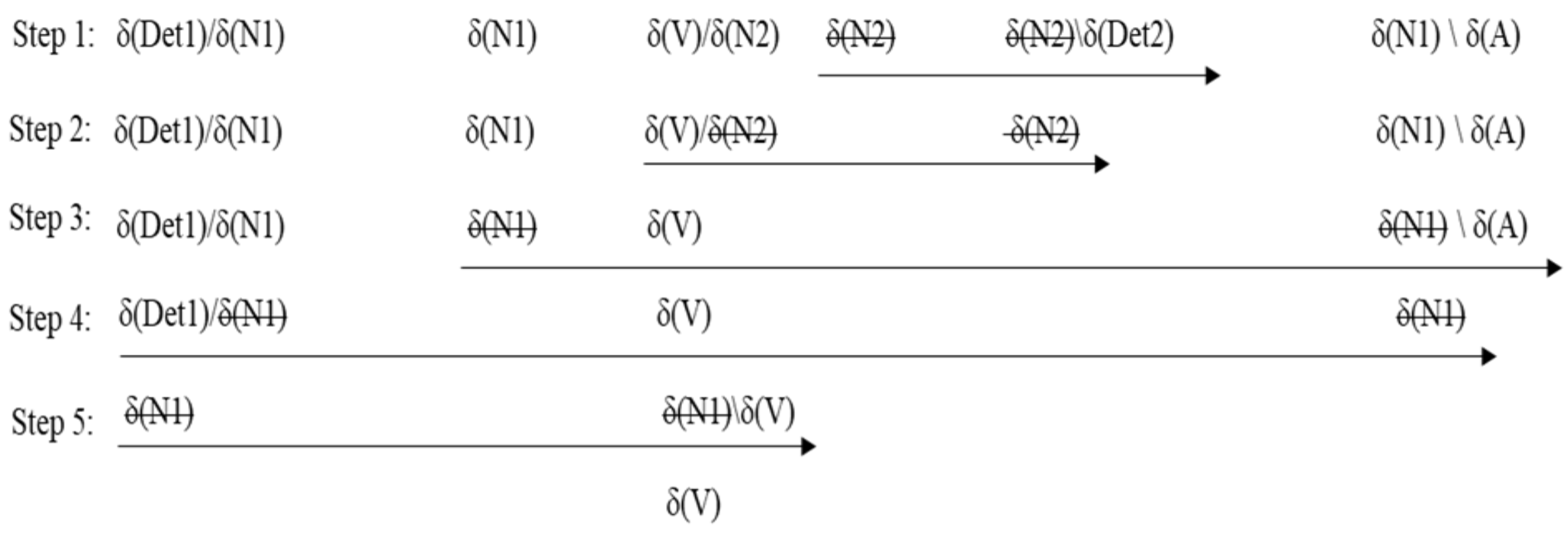

Fig 40. A unified representation

In all, the derivations formulated for the three illustrative cases above show how the conversions, namely PSG->CG, DG->CG, CG->DG and CG->PSG (not necessarily always in that order), can establish the desired equivalence of representations in those formalisms. Therefore, establishing PSG->CG, DG->CG, CG->DG and CG->PSG is tantamount to establishing PSG≡CG≡DG in their representational descriptions of natural language constructions.

## 7. Implications and Conclusion

This paper has attempted to show that the understanding of constituency with respect to re-writing rules, categorial relations and dependency relations can be integrated into one system of representation. This unified representation provides evidence for the possibility of approaching discontinuity in natural language by depicting the manner in which the three grammar formalisms PSG, CG and DG, which are usually regarded as sort of non-unifiable approaches, can account for the notion of constituency in a similar manner. We have demonstrated how a flexible account of functor-argument relations can be decoded from the rigid constituents of phrase structures and how in turn these functor-argument (categorial) relations can also be formulated in terms of dependency relations. Hence, the PSG rules in trees could be redrawn in terms of the CG formula which in turn could be rewritten in terms of the DG functions. This can have far-reaching implications for theories of natural language since most current linguistic theories do adopt and subscribe to constituency-based analyses, although specific treatments of particular phenomena such as labelling phrases may differ. But one emerging conclusion is that not all aspects of natural language (especially syntax) can be accounted for by binary branching and headed rules (see Müller 2013). The unified system of representation for continuous and discontinuous structures cuts across and in fact (somewhat) neutralizes the traditional distinction between derivational theories (as in Chomsky 1995) and constraint-based formalisms (such as Lexical Functional Grammar). That is mainly because certain kinds of discontinuity have been hard to handle within derivational theories, while the segregation of phrase structure from functional-semantic specifications within constraint-based formalisms has made it feasible to capture such cases of discontinuity (since parts of a discontinuous expression within phrase structure are directly translated into functional specifications). Besides, both types of formalisms have to define their derivations or constraints on the structural organization of linguistic structures anyway.

Given that representations of linguistic structures in PSG, CG and DG can be recast in a single system of representation, it appears that the tensions between constituency-based representations and non-constituency representations are more of a theoretical artefact than a linguistic reality. Thus, this unification of the three approaches can serve a general linguistic theory better, in that it can account for both continuity and discontinuity, not in separate ways with different assumptions, but in a single descriptive system of representation. We propose that this work could also hint at a possibility that there could be a single representation for continuous and discontinuous structures in our cognitive system as continuous and discontinuous structures observed in natural languages have to be handled by the neurocognitive system anyway. We leave it open as a matter of further investigation, though.